\pdfoutput=1

\documentclass[11pt]{article}

\usepackage[preprint]{acl}

\usepackage{times}
\usepackage{latexsym}

\usepackage[T1]{fontenc}

\usepackage[utf8]{inputenc}

\usepackage{microtype}

\usepackage{inconsolata}

\usepackage{amsmath}
\usepackage{amssymb}
\usepackage{graphicx}
\usepackage{multirow}
\usepackage{threeparttable}
\usepackage{booktabs}
\usepackage{colortbl}

\definecolor{mygray}{gray}{.9}
\definecolor{mypink}{rgb}{.99,.91,.95}
\definecolor{mycyan}{cmyk}{.3,0,0,0}

\title{Are AI-Generated Text Detectors Robust to Adversarial Perturbations?}

\author{Guanhua Huang$^{1*}$, Yuchen Zhang$^{2 \dagger}$, Zhe Li$^{2}$, Yongjian You$^{2}$, \\
{\bf Mingze Wang$^{3}$} \and {\bf Zhouwang Yang$^{1 \dagger}$} \\
        $^{1}$University of Science and Technology of China $^{2}$Bytedance $^{3}$Peking University \\
        \texttt{guanhuahuang@mail.ustc.edu.cn} \\
        \texttt{\{zhangyuchen.zyc, lizhe.2023, youyongjian.cc\}@bytedance.com} \\
        \texttt{mingzewang@stu.pku.edu.cn, yangzw@ustc.edu.cn}
        }

\begin{document}
\maketitle
\renewcommand{\thefootnote}{\fnsymbol{footnote}}
\footnotetext[1]{Work done during ByteDance Research internship.}
\footnotetext[2]{Corresponding author.}
\renewcommand{\thefootnote}{\arabic{footnote}}
\begin{abstract}
The widespread use of large language models (LLMs) has sparked concerns about the potential misuse of AI-generated text, as these models can produce content that closely resembles human-generated text. Current detectors for AI-generated text (AIGT) lack robustness against adversarial perturbations, with even minor changes in characters or words causing a reversal in distinguishing between human-created and AI-generated text. This paper investigates the robustness of existing AIGT detection methods and introduces a novel detector, the Siamese Calibrated Reconstruction Network (SCRN). The SCRN employs a reconstruction network to add and remove noise from text, extracting a semantic representation that is robust to local perturbations. We also propose a siamese calibration technique to train the model to make equally confident predictions under different noise, which improves the model's robustness against adversarial perturbations. Experiments on four publicly available datasets show that the SCRN outperforms all baseline methods, achieving 6.5\%-18.25\% absolute accuracy improvement over the best baseline method under adversarial attacks. Moreover, it exhibits superior generalizability in cross-domain, cross-genre, and mixed-source scenarios. The code is available at \url{https://github.com/CarlanLark/Robust-AIGC-Detector}.
\end{abstract}

\section{Introduction}

Large Language Models (LLMs) such as GPT-4 \citep{achiam2023gpt} have shown great promise in producing text that closely mimics human language \citep{wang-etal-2023-element, chiang-lee-2023-large, park2023generative}. However, concerns about the misuse of AI-generated text (AIGT) have arisen in various areas, including the spread of fake news \citep{hanley2023machine}, academic dishonesty \citep{perkins2023game}, and gender bias \citep{wan-etal-2023-kelly}. To tackle these issues, various AIGT detection methods have been developed, using statistical features from language models and text features from different model architectures and training approaches \citep{solaiman2019release, gehrmann-etal-2019-gltr, mitchell-etal-2023-detectgpt, guo2023close}.

\begin{figure}[t]
\centering 
\includegraphics[width = 0.5\textwidth]{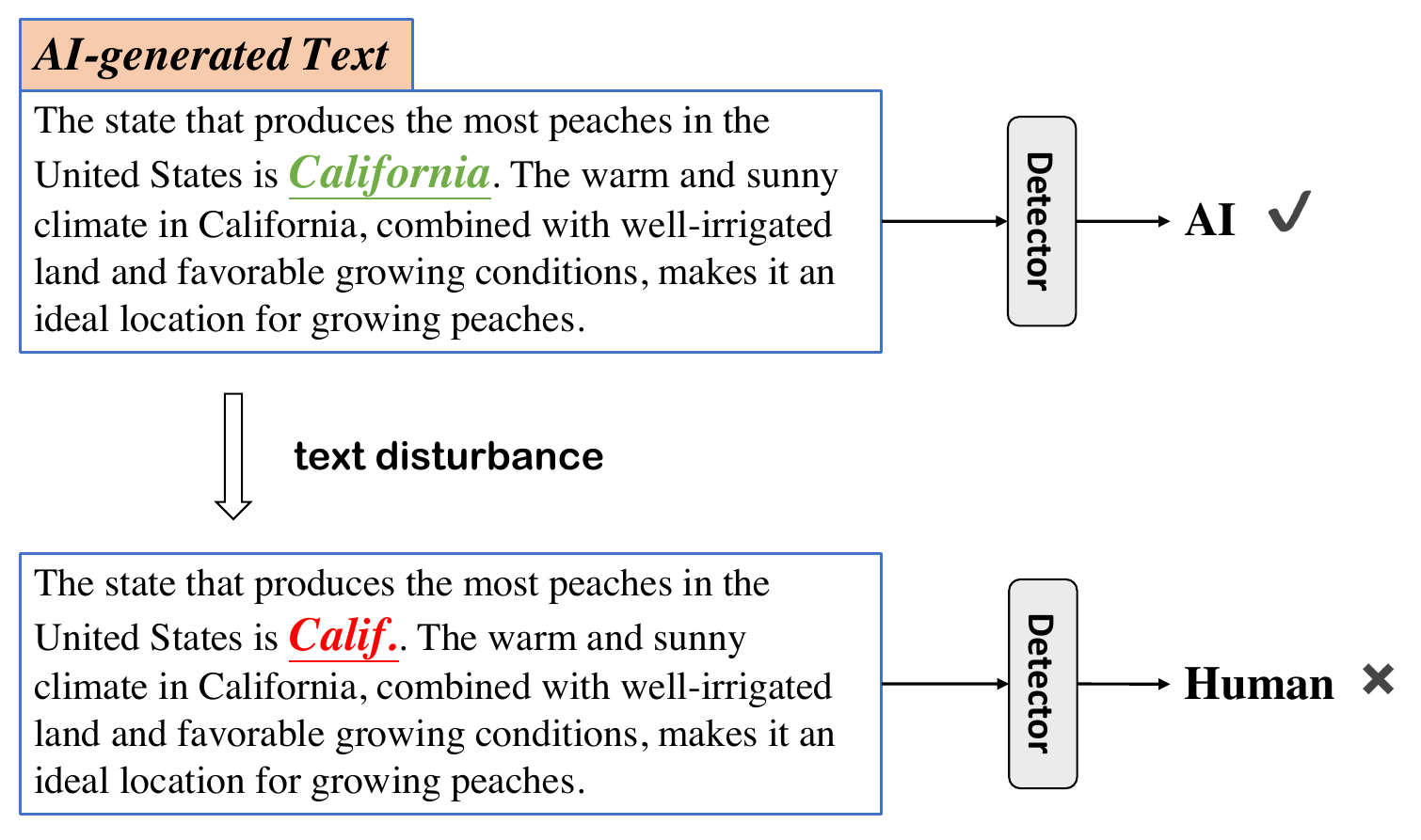} 
\caption{An example of adversarial perturbation to a RoBERTa-based AIGT detector.} 
\label{fig:motivation} 
\end{figure}

Current AI-generated text (AIGT) detectors can effectively identify AI-generated text but struggle with minor adversarial perturbations, such as word substitutions or character swapping \citep{peng-etal-2023-hidding, cai2023evade}. Small changes that do not change the original text's meaning can cause these detectors to fail.
Figure \ref{fig:motivation} shows a concrete example: a RoBERTa-based AIGT detector can be fooled into classifying AI-generated text as human-written by simply abbreviating "California" to "Calif." This example underscores the limitations of relying solely on token-level features. Therefore, developing robust AIGT detection methods that rely on high-level features is crucial to counteract adversarial perturbation attacks.

To address these challenges, we introduce the Siamese Calibrated Reconstruction Network (SCRN), which consists of an encoder, a reconstruction network, and a classification head. The model first converts input texts into token representations, then introduces random Gaussian noise to simulate a perturbation attack. The reconstruction network, acting as a denoising auto-encoder \citep{vincent2008extracting}, aims to remove this noise and recover the original representations. The classification head processes these denoised features to produce the final result. During training, we optimize both classification and reconstruction losses, encouraging the model to learn representations that are resilient to random input perturbations.

Empirically, we observe that a model trained for robustness against random perturbations may not necessarily be robust against adversarial perturbations. To address this issue, we introduce a training technique called siamese calibration. During training, the model generates two classification results using two independent sets of random noise. The training procedure aims to minimize the symmetric Kullback–Leibler (KL) divergence between the two output probability distributions. Since KL divergence is sensitive to changes in probabilities at all confidence levels, the model can incur a significant loss even when it makes consistently correct predictions but with varying confidence levels due to different noise. This stronger constraint forces the model to make equally confident predictions regardless of the noise. In experiments, we find that this approach encourages the model to rely more on high-level contextual features, thereby significantly enhancing its robustness against adversarial attacks.

Our contributions are as follows:

(1) We introduce a reconstruction network that enhances the model's robustness by promoting the learning of resilient representations under token-level perturbations.

(2) We propose a siamese calibration technique that trains the model to make predictions with consistent confidence levels for various random perturbations, which improves its robustness against adversarial attacks.

(3) We establish a comprehensive benchmark for assessing the robustness of AIGT detection methods against a range of adversarial perturbations, including word-level and character-level substitution, deletion, insertion, and swapping. This benchmark encompasses a wide variety of detectors, such as metric-based and model-based detectors, trained using different methods. We evaluate these detectors on four publicly available datasets to test their robustness in in-domain, cross-domain, cross-genre, and mixed-source scenarios.

(4) Our experiments on the benchmark show that SCRN significantly outperforms all baselines in terms of robustness, achieving higher accuracy under adversarial perturbation attacks. Specifically, our method improves over the best baseline method by 11.25, 18.25, 14.5, and 15.75 absolute points 
of accuracy under attack in in-domain, cross-domain, cross-genre, and mixed-source scenarios, respectively.

\section{Related Work}

In recent years, Large Language Models (LLMs) like GPT-2 \citep{Radford2019LanguageMA} and GPT-3 \citep{Brown2020LanguageMA} have shown impressive performance in various natural language generation tasks \citep{kamalloo-etal-2023-evaluating, wang-etal-2023-element, cheng2023cl, chiang-lee-2023-large, park2023generative, qin2023toolllm}. However, the advent of more advanced models such as GPT-4 \citep{achiam2023gpt} has raised concerns about the potential misuse of AI-generated texts (AIGT) in areas like fake news \citep{hanley2023machine, zhou2023synthetic}, academic cheating \citep{perkins2023game, foltynek2023enai}, and ingroup bias \citep{wan-etal-2023-kelly, gallegos2023bias}. This highlights the importance of robust detection mechanisms to ensure the security and reliability of applications using LLMs.

To differentiate between human-written and AI-generated texts, various AIGT detection methods have been developed \citep{gehrmann-etal-2019-gltr, ippolito-etal-2020-automatic, uchendu-etal-2021-turingbench-benchmark, guo2023close, mitchell-etal-2023-detectgpt}. These methods fall into two categories: metric-based and model-based. Metric-based methods use a language model to generate scores for the text and create statistical features from them, such as probability score \citep{solaiman2019release}, rank score \citep{mitchell-etal-2023-detectgpt}, and entropy score \cite{gehrmann-etal-2019-gltr}. Model-based methods, on the other hand, employ neural network architectures and supervised learning to train detectors end-to-end using labeled text. For example, OpenAI trained a RoBERTa model to detect GPT-2-generated text \citep{solaiman2019release}, while \citep{guo2023close} developed a ChatGPT detector based on question-answer text from various domains.

However, research shows that both metric-based and model-based detectors are susceptible to adversarial perturbations \citep{cai2023evade, peng-etal-2023-hidding}, such as synonym replacement \citep{ren-etal-2019-generating, Jin_Jin_Zhou_Szolovits_2020}. This means that minor word changes can lead to incorrect classification of AI-generated texts in various fields, including news, education, and finance.\citep{peng-etal-2023-hidding} While robust methods have been introduced in related areas like sentiment analysis \citep{wang-etal-2023-rmlm} and speech recognition \cite{cheng2023ml}, a comprehensive analysis of the resilience of AIGT detectors against adversarial perturbations remains lacking. In this work, we aim to explore the robustness of existing detectors against adversarial perturbations, both in-domain and out-of-domain.

\section{Siamese Calibrated Reconstruction Network}

\begin{figure*}[!t] 
\centering 
\includegraphics[width=1.0\textwidth]{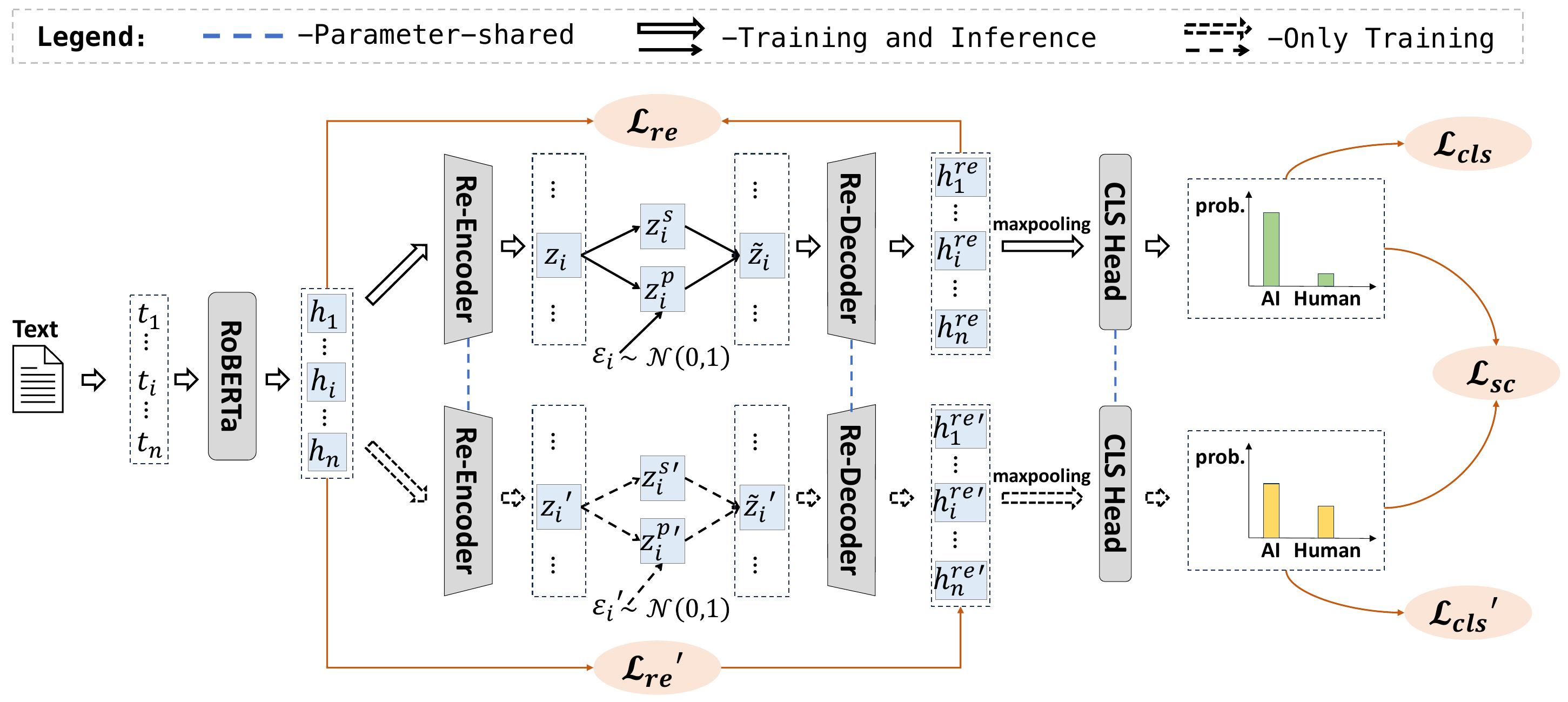} 
\caption{The architecture of SCRN. The input is first encoded by a pre-trained RoBERTa encoder. Then the representation is mapped to a lower-dimensional space by the Re-Encoder to construct the semantic term and the perturbation term, based on which the representation is reconstructed by the Re-Decoder. The denoised representation is used to predict class distributions. Finally, a discrepancy loss is minimized to calibrate the class distributions of two parameter-shared branches.} 
\label{fig:framework} 
\end{figure*}

\subsection{Model Architecture}

\paragraph{Encoder}

Given a dataset $\mathcal{D} = \{(x, y)\}$, where $x = (w_1, w_2, ..., w_n)$ is the input text with $n$ tokens and $y$ denotes the binary label (human or AI), we use a pre-trained RoBERTa \citep{liu2019roberta} as the encoder to obtain the text representation:
\begin{equation*}
[h_1, h_2, ..., h_n] = {\rm RoBERTa}(w_1, w_2, ..., w_n)
\end{equation*}
where $h_i \in \mathbb{R}^{d}$ are the encoded tokens.

\paragraph{Reconstruction Network}

To make the detector more robust to text perturbations, we simulate an actual perturbation by adding random noise to each $h_i$, then utilize a reconstruction network to remove the noise. To inject noise, we split the token representation into a semantic term and a perturbation term, where the former retains the semantic meaning and the latter contains noise. This approach is inspired by \cite{john-etal-2019-disentangled}, which separates the text representation into semantic and style terms to control text style.

Specifically, for the $i$-th token representation $h_i \in \mathbb{R}^{d}$, we use an MLP encoder to map it to a lower-dimensional latent space:
\begin{align*}
z_i = {\rm MLP}^{(enc)}(h_i)
\end{align*}
where $z_i \in \mathbb{R}^{d_z}$ is the $i$-th latent representation. We define a semantic term $z_i^{(s)}$ and a perturbation term $z_i^{(p)}$ based on $z_i$ as follows:
\begin{align*}
z_i^{(s)} &= W^{(s)} z_i + b^{(s)} \\
z_i^{(p)} &= W^{(p)} z_i + b^{(p)}
\end{align*}
Here, $W^{(s)} \in \mathbb{R}^{d_z \times d_z}, W^{(p)} \in \mathbb{R}^{d_z \times 1}, b^{(s)} \in \mathbb{R}^{d_z}, b^{(p)} \in \mathbb{R}$ are trainable parameters.

Then we combine the two terms to define a noisy latent representation:
\begin{align*}
    \tilde{z}_i = z_i^{(s)} + \epsilon \cdot z_i^{(p)} \cdot \mathbb{I}
\end{align*}
where $\epsilon \sim \mathcal{N}(0, 1)$ is a standard Gaussian noise and $\mathbb{I} \in \mathbb{R}^{d_z}$ is a scalar vector with each entry equal to 1.

To control the numerical scale of latent representations, we introduce a regularization term:
\begin{equation*}
\mathcal{L}_{\rm reg}(x) \!= \frac{1}{n} \sum_{i=1}^n \!\Big(\|z_i^{(s)}\|_2^2 \!+ {|z_i^{(p)}|}^2 \!- \alpha \cdot \log(|z_{i}^{(p)}|) \!\Big)
\end{equation*}
Here, the first two terms serve to penalize the large numerical scale of the latent representations, while the third term aims to prevent the noise scale from decreasing to zero. The hyperparameter $\alpha$ allows for adjusting the degree of penalty applied.

Finally, an MLP decoder is applied to reconstruct the original token representation based on the noisy latents:
\begin{align*}
    h_i^{(re)} = {\rm MLP}^{(dec)}(\tilde{z}_i)
\end{align*}
where $h_i^{(re)} \in \mathbb{R}^{d}$ represents the reconstructed $i$-th token representation. The reconstruction error is the mean square error between the reconstructed representation and the original one:
\begin{align*}
    \mathcal{L}_{\rm mse}(x) = \frac{1}{n} \sum_{i=1}^n ||h_i^{(re)} - h_i||_2^2
\end{align*}
The final reconstruction loss is the sum of the reconstruction error and the regularization term, where $\beta$ is a hyperparameter.
\begin{align*}
    \mathcal{L}_{\rm re} = - \frac{1}{|\mathcal{D}|} \sum_{x \in \mathcal{D}} (\mathcal{L}_{\rm mse}(x) + \beta \cdot \mathcal{L}_{\rm reg}(x))
\end{align*}

\paragraph{Classification Head}
After obtaining the reconstructed token representations, we use a max pooling layer to extract the feature of the final classification:
\begin{align*}
h^{(cls)} = \text{MaxPooling}([h_1^{(re)}, ..., h_n^{(re)}]).
\end{align*}
Then we predict the label $\hat{y}$ using a MLP classifier. The classification loss is the standard cross-entropy loss, denoted by $\mathcal{L}_{\rm cls}$.

\subsection{Siamese Calibration}

The reconstruction loss helps the detector learn robust representations against random perturbations, but it does not guarantee robustness against targeted adversarial perturbations. Empirically, we find that the model, optimized with both reconstruction and classification losses, remains vulnerable to adversarial attacks.

To address this issue, we propose a siamese calibration training strategy. This strategy aims to minimize the symmetric Kullback-Leibler (KL) divergence between the outputs of two inference branches, each subjected to independent random noises, given the same input. Specifically, let $P(x,\epsilon)$ be the predicted class distribution for input $x$ with noise $\epsilon$. The symmetric divergence is then defined as the average of $D_{\rm KL}(P(x,\epsilon)||P(x,\epsilon'))$ and $D_{\rm KL}(P(x,\epsilon')||P(x,\epsilon))$, where $\epsilon$ and $\epsilon'$ are independent copies of random noise, and $D_{\rm KL}$ represents the Kullback–Leibler (KL) divergence. Unlike the cross-entropy loss, which is negligible for correct predictions with high confidence, the symmetric divergence is sensitive to all confidence levels. 
For instance, consider the correct label is AI. If the first branch predicts AI with a probability of 0.99 and the second branch predicts AI with a probability of $1-\delta$ perturbated by a different random noise, where $\delta<0.01$, the cross-entropy loss for both branches is less than 0.01, providing little incentive for further optimization. However, the symmetric divergence between these two distributions can approach infinity as $\delta\rightarrow 0$. 
We define the \emph{siamese calibration loss}, denoted as $\mathcal{L}_{\rm sc}$, as the symmetric divergence across all training inputs $x\in D$. This loss specifically penalizes inconsistent confidence levels, imposing a stricter requirement than the cross-entropy loss. By minimizing this loss, the detector is encouraged to focus on high-level contextual features that are less susceptible to token perturbations.

For training the SCRN model, we make a weighted summing over the above three losses:
\begin{align*}
    \mathcal{L}_{\rm all} = \lambda_1 (\mathcal{L}_{\rm cls} + \mathcal{L}_{\rm cls}') + \lambda_2 (\mathcal{L}_{\rm re} + \mathcal{L}_{\rm re}') + \lambda_3 \mathcal{L}_{\rm sc}
\end{align*}
where $\mathcal{L}_{\rm cls}, \mathcal{L}_{\rm cls}'$ are classification losses of two branches, $\mathcal{L}_{\rm re}, \mathcal{L}_{\rm re}'$ are reconstruction losses, and $\mathcal{L}_{\rm sc}$ is siamese calibration loss. $ \lambda_1, \lambda_2, \lambda_3$ are hyperparameters.

Siamese calibration is only applicable during training. During inference, the detector generates a single copy of random noise and produces a single prediction. Training with siamese calibration significantly enhances the model's prediction consistency during inference. Analysis in Appendix \ref{sec:Inference Fluctuation Analysis} shows that the fluctuation in inference robustness between two independent branches becomes negligible after implementing siamese calibration.

\section{Experiments}
\subsection{Experimental Setup}

\begin{table*}[!h]
\centering
\scalebox{0.8}{
\begin{tabular}{lccccc}
\toprule
Scenarios    & Train Set  & Test Set  & Num. Train & Num. Test & Num. Attack \\
\midrule
In-domain    & \texttt{HC3 train} & \texttt{HC3 test}    & 76,905     & 8,544     & 400         \\
Cross-domain & \texttt{HC3 train} & \texttt{TruthfulQA}  & 76,905     & 1,634     & 400         \\
Cross-genre  & \texttt{HC3 train} & \texttt{GhostBuster} & 76,905     & 6,000     & 400  \\
Mixed-source & \texttt{SeqXGPT-Bench train} & \texttt{SeqXGPT-Bench test} & 10,800 & 1,200 & 400 \\
\bottomrule
\end{tabular}
}
\caption{Statistics of datasets for different AIGT detection scenarios.}
\label{tab:dataset-statistics}
\end{table*}

\paragraph{Datasets}
To assess the robustness of AIGT detectors to adversarial perturbations, we conduct experiments on four public datasets: \textbf{\texttt{HC3}}\citep{guo2023close}, \textbf{\texttt{TruthfulQA}}\citep{he2023mgtbench}, \textbf{\texttt{Ghostbuster}}\citep{verma2023ghostbuster}, and \textbf{\texttt{SeqXGPT-Bench}}\citep{wang-etal-2023-seqxgpt}. These datasets encompass a wide range of AIGT scenarios, including in-domain, cross-domain, cross-genre, and mixed-source. The datasets and number of samples used for training and evaluating AIGT detectors are listed in Table \ref{tab:dataset-statistics}. For dataset details, see Appendix \ref{sec:Datasets}.

\paragraph{Metrics}
Following the approach of \cite{wang-etal-2023-rmlm}, we assess model accuracy and robustness using four metrics: (1) Original Accuracy (\textbf{OA}\% $\uparrow$) measures the model's raw accuracy without adversarial perturbations. (2) Accuracy Under Attack (\textbf{AUA}\% $\uparrow$) quantifies the model's accuracy on adversarially perturbed text. (3) Attack Success Rate (\textbf{ASR}\% $\downarrow$) indicates the percentage of test samples successfully fooled by the attacker. (4) Average Number of Queries (\textbf{ANQ} $\uparrow$) represents the average number of adversarial attack queries on test samples, with higher values indicating a more robust model. The symbols $\uparrow$ and $\downarrow$ denote that higher and lower values are better, respectively. Additionally, we report traditional \textbf{Precision}, \textbf{Recall}, and \textbf{micro-F1} scores when no attack is performed.

\paragraph{Adversarial Perturbations}
In our experiments, adversarial perturbations are conducted by three attack methods: \textbf{PWWS} \citep{ren-etal-2019-generating}, \textbf{Deep-Word-Bug} \citep{gao-etal-2018-deepwordbug}, and~\textbf{Deep-Word-Bug} \citep{gao-etal-2018-deepwordbug}. These methods involve various types of adversarial perturbations including character-level and word-level substitution, deletion, and insertion. More details can be found in Appendix \ref{sec:Adversarial Perturbations}

\paragraph{Baselines} To create a comprehensive benchmark, we do our best to select a wide range of detector methods. These include metric-based detectors such as \textbf{Log-Likelihood}\citep{solaiman2019release}, \textbf{Log-Rank}\citep{mitchell-etal-2023-detectgpt}, \textbf{Entropy}\citep{gehrmann-etal-2019-gltr}, \textbf{GLTR}\citep{gehrmann-etal-2019-gltr}, and \textbf{SeqXGPT}\cite{wang-etal-2023-seqxgpt}, along with model-based detectors like \textbf{Bert}\citep{devlin-etal-2019-bert}, \textbf{Roberta}\citep{liu2019roberta}, \textbf{Deberta}\citep{he2020deberta}, and \textbf{ChatGPT-Detector}\citep{guo2023close}. We also implemented recent methods for enhancing classification robustness in other NLP tasks, such as \textbf{Flooding}\citep{pmlr-v119-ishida20a}, \textbf{RDrop}\citep{wu2021r}, \textbf{RanMASK}\citep{zeng-etal-2023-certified}, and \textbf{RMLM} \citep{wang-etal-2023-rmlm}. Details can be found in Appendix \ref{sec:Baselines}.

\paragraph{Experiment Settings}
To ensure a fair comparison of the AIGT detectors, all models were trained on 8 * 32GB NVIDIA V100 GPUs and evaluated on a single 32GB NVIDIA V100 GPU, using the same environment. We utilized the base versions of pre-trained Bert, RoBERTa, and DeBERTa models as employed in the compared detectors. The implemented methods were based on their officially released code, and for the ChatGPT-Detector, we utilized the provided model weight. The hyperparameters of SCRN and more information can be found in Appendix \ref{sec:Experiment Settings}.

\subsection{In-domain Robustness} \label{sec:In-domain Robustness}

\begin{table}[!h]
\centering
\scalebox{0.7}{
\begin{tabular}{lccccc}
\toprule
Methods & P.(AI) & R.(AI) & P.(H.) & R.(H.) & F.(Overall) \\ \midrule
Log-Likelihood   & 95.43 & 95.61 & 97.98 & 97.90 & 97.18 \\
Log-Rank         & 96.30 & 95.87 & 98.11 & 98.31 & 97.54 \\
Entropy          & 89.90 & 87.69 & 94.41 & 95.47 & 93.00 \\
GLTR             & 96.84 & 95.76 & 98.06 & 98.57 & 97.68 \\
SeqXGPT          & 97.92 & \textbf{100.0} & \textbf{100.0} & 99.03 & 99.33 \\
Bert             & 98.28 & \textbf{100.0} & \textbf{100.0} & 99.20 & 99.45 \\
Roberta          & 99.45 & \textbf{100.0} & \textbf{100.0} & 99.74 & 99.80 \\
Deberta          & 99.74 & \textbf{100.0} & \textbf{100.0} & 99.88 & 99.92 \\
ChatGPT-Detector & 99.03 & 99.15 & 99.61 & 99.56 & 99.58 \\
Flooding         & 99.67 & \textbf{100.0} & \textbf{100.0} & 99.85 & 99.89 \\
RDrop            & 99.67 & 99.96 & 99.98 & 99.85 & 99.88 \\
RanMASK          & 87.73 & \textbf{100.0} & \textbf{100.0} & 93.58 & 95.67 \\
RMLM             & 96.15 & \textbf{100.0} & \textbf{100.0} & 96.00 & 98.00 \\
\rowcolor{mygray}
SCRN             & \textbf{99.78} & \textbf{100.0} & \textbf{100.0} & \textbf{99.90} & \textbf{99.93} \\ \bottomrule
\end{tabular}
}
\caption{Results of \textbf{in-domain} AIGT detection \textbf{without attack}. P.(AI) and R.(AI) denote Precision and Recall for AI-generated text as positive samples, while P.(H.) and R.(H.) indicate Precision and Recall for human-created text as positive samples.}
\label{tab:in-domain-without-attack}
\end{table}

\begin{table*}[!h]
\centering
\scalebox{0.7}{
\begin{tabular}{ll|cccc|cccc|cccc}
\toprule
 &
   &
  \multicolumn{4}{c|}{AI $\rightarrow$ Human} &
  \multicolumn{4}{c|}{Human $\rightarrow$ AI} &
  \multicolumn{4}{c}{Overall} \\
& Methods & OA $\uparrow$ & AUA $\uparrow$ & ASR $\downarrow$ & ANQ $\uparrow$ & OA $\uparrow$ & AUA $\uparrow$ & ASR $\downarrow$ & ANQ $\uparrow$ & OA $\uparrow$ & AUA $\uparrow$ & ASR $\downarrow$ & ANQ $\uparrow$ \\
\midrule
\multirow{13}{*}{\rotatebox{90}{PWWS}}   
& Log-Likelihood & 96.00  & 0.00  & 100.00 & 957.42   & \textbf{100.00} & 99.00  & 1.00  & 1223.54 & 98.00  & 49.50 & 49.49 & 1090.48  \\
& Log-Rank       & 96.50  & 0.00  & 100.00 & 974.20   & 99.00  & 98.50  & 0.51  & 1233.61 & 97.75  & 49.25 & 49.62 & 1103.90  \\
& Entropy        & 86.00  & 0.00  & 100.00 & 962.97   & 95.00  & 79.00  & 16.84 & 1112.75 & 90.50  & 39.50 & 56.35 & 1037.86  \\
& GLTR           & 95.00  & 0.00  & 100.00 & 986.10   & 99.00  & 83.50  & 15.66 & 1136.52 & 97.00  & 41.75 & 56.96 & 1061.31  \\
& SeqXGPT        & 99.50  & 11.00 & 88.94  & 1368.07  & \textbf{100.00} & \textbf{100.00} & \textbf{0.00}  & 1224.88 & 99.75  & 55.50 & 44.36 & 1296.47  \\
& BERT           & \textbf{100.00} & 1.50  & 98.50  & 1070.74  & 99.50  & 98.50  & 1.01  & 1211.18 & 99.75  & 50.00 & 49.87 & 1140.96  \\
& RoBERTa        & \textbf{100.00} & 38.50 & 61.50  & 1332.60  & \textbf{100.00} & 99.50  & 0.50  & 1223.76 & \textbf{100.00} & 69.00 & 31.00 & 1278.18  \\
& DeBERTa        & \textbf{100.00} & 3.00  & 97.00  & 1170.89  & \textbf{100.00} & 99.50  & 0.50  & 1223.53 & \textbf{100.00} & 51.25 & 48.75 & 1197.21  \\
& ChatGPT-Detector      & 98.00  & 0.00  & 100.00 & 1074.98  & \textbf{100.00} & 99.50  & 0.50  & 1224.60 & 99.00  & 49.75 & 49.75 & 1149.79  \\
& Flooding       & \textbf{100.00} & 23.00 & 77.00  & 1422.60  & \textbf{100.00} & \textbf{100.00} & \textbf{0.00 } & 1225.00 & \textbf{100.00} & 61.50 & 38.50 & 1323.81  \\
& RDrop & 99.50  & 67.50 & 32.16 & 1585.15 & \textbf{100.00} & \textbf{100.00} & \textbf{0.00} & \textbf{1225.12} & 99.75  & 83.75 & 16.04 & 1405.08 \\
& RanMASK        & \textbf{100.00} & 50.00 & 50.00  & 1562.84  & \textbf{100.00} & \textbf{100.00} & \textbf{0.00}  & 1245.97 & \textbf{100.00} & 75.00 & 25.00 & 1404.40  \\
& RMLM           & \textbf{100.00} & 73.50 & 26.50  & 1561.35  & \textbf{100.00} & 98.50  & 1.50  & 1216.52 & \textbf{100.00} & 86.00 & 14.00 & 1388.94  \\
\rowcolor{mygray}
& SCRN           & \textbf{100.00} & \textbf{94.50} & \textbf{5.50}   & \textbf{1665.53}  & \textbf{100.00} & \textbf{100.00} & \textbf{0.00}  & 1225.02 & \textbf{100.00} & \textbf{97.25} & \textbf{2.75}  & \textbf{1445.28}  \\
\midrule
\multirow{13}{*}{\rotatebox{90}{Deep-Word-Bug}} 
& Log-Likelihood & 96.00 & 0.00 & 100.00 & 109.20 & \textbf{100.00} & \textbf{100.00} & \textbf{0.00} & 306.24 & 98.00 & 50.00 & 48.98 & 207.72 \\
& Log-Rank       & 96.50  & 0.00  & 100.00 & 110.93   & 99.00  & 99.00  & \textbf{0.00}  & 308.03  & 97.75  & 49.50 & 49.36 & 209.48   \\
& Entropy        & 86.00  & 0.00  & 100.00 & 108.82   & 95.00  & 92.50  & 2.63  & 295.70  & 90.50  & 46.25 & 48.90 & 202.26   \\
& GLTR           & 95.00  & 0.00  & 100.00 & 114.79   & 99.00  & 98.50  & 0.51  & 306.09  & 97.00  & 49.25 & 49.23 & 210.44   \\
& SeqXGPT        & 99.50  & 8.00  & 91.96  & 139.96   & \textbf{100.00} & \textbf{100.00} & \textbf{0.00}  & 306.16  & 99.75  & 54.00 & 45.86 & 223.06   \\
& BERT           & \textbf{100.00} & 12.50 & 87.50  & 152.30   & 99.50  & 98.50  & 1.01  & 282.92  & 99.75  & 55.50 & 44.36 & 217.61   \\
& RoBERTa        & \textbf{100.00} & 53.00 & 47.00  & 293.59   & \textbf{100.00} & \textbf{100.00} & \textbf{0.00}  & 302.58  & \textbf{100.00} & 76.50 & 23.50 & 298.08   \\
& DeBERTa        & \textbf{100.00} & 32.00 & 68.00  & 171.36   & \textbf{100.00} & \textbf{100.00} & \textbf{0.00}  & 295.56  & \textbf{100.00} & 66.00 & 34.00 & 233.46   \\
& ChatGPT-Detector      & 98.00  & 13.50 & 86.22  & 161.07   & \textbf{100.00} & \textbf{100.00} & \textbf{0.00}  & 301.62  & 99.00  & 56.75 & 42.68 & 231.34   \\
& Flooding       & \textbf{100.00} & 35.00 & 65.00  & 175.48   & \textbf{100.00} & \textbf{100.00} & \textbf{0.00}  & 275.71  & \textbf{100.00} & 67.50 & 32.50 & 225.59   \\
& RDrop & 99.50  & 60.50 & 39.20 & 367.74  & \textbf{100.00} & \textbf{100.00} & \textbf{0.00} & 306.09  & 99.75  & 80.25 & 19.55 & 336.92  \\
& RanMASK        & \textbf{100.00} & 59.00 & 41.00  & 332.98   & \textbf{100.00} & \textbf{100.00} & \textbf{0.00}  & \textbf{315.78}  & \textbf{100.00} & 79.50 & 20.50 & 324.38   \\
& RMLM           & \textbf{100.00} & 66.00 & 34.00  & 377.24   & \textbf{100.00} & \textbf{100.00} & \textbf{0.00}  & 308.91  & \textbf{100.00} & 83.00 & 17.00 & 343.08   \\
\rowcolor{mygray}
& SCRN           & \textbf{100.00} & \textbf{87.50} & \textbf{12.50}  & \textbf{437.50}   & \textbf{100.00} & \textbf{100.00} & \textbf{0.00}  & 305.93  & \textbf{100.00} & \textbf{93.75} & \textbf{6.25}  & \textbf{371.72}   \\
\midrule
\multirow{13}{*}{\rotatebox{90}{Pruthi}} 
& Log-Likelihood & 96.00  & 1.00  & 98.96  & 9000.48  & \textbf{100.00} & \textbf{100.00} & \textbf{0.00}  & 9519.34 & 98.00  & 50.50 & 48.47 & 9259.91  \\
& Log-Rank       & 96.50  & 1.00  & 98.96  & 10084.75 & 99.00  & 99.00  & \textbf{0.00}  & 9557.18 & 97.75  & 50.00 & 48.85 & 9820.96  \\
& Entropy        & 86.00  & 0.00  & 100.00 & 7920.77  & 95.00  & 90.50  & 4.74  & 9085.92 & 90.50  & 45.25 & 50.00 & 8503.35  \\
& GLTR           & 95.00  & 1.00  & 98.95  & 10321.49 & 99.00  & 95.00  & 4.04  & 9507.22 & 97.00  & 48.00 & 50.52 & 9914.36  \\
& SeqXGPT        & 99.50  & 1.00  & 98.99  & 10505.46 & \textbf{100.00} & \textbf{100.00} & \textbf{0.00}  & 9609.78 & 99.75  & 50.50 & 49.37 & 10057.62 \\
& BERT           & \textbf{100.00} & 21.00 & 79.00  & 17460.63 & 99.50  & 98.00  & 1.51  & 9502.22 & 99.75  & 59.50 & 40.35 & 13481.42 \\
& RoBERTa        & \textbf{100.00} & 57.00 & 43.00  & 20431.19 & \textbf{100.00} & \textbf{100.00} & \textbf{0.00}  & 9561.62 & \textbf{100.00} & 78.50 & 21.50 & 14996.40 \\
& DeBERTa        & \textbf{100.00} & 34.50 & 65.50  & 17338.08 & \textbf{100.00} & 99.50  & 0.50  & 9606.24 & \textbf{100.00} & 67.00 & 33.00 & 13472.16 \\
& ChatGPT-Detector      & 98.00  & 36.50 & 62.76  & 18563.57 & \textbf{100.00} & 99.50  & 0.50  & 9591.88 & 99.00  & 68.00 & 31.31 & 14077.72 \\
& Flooding       & \textbf{100.00} & 68.50 & 31.50  & 20823.59 & \textbf{100.00} & \textbf{100.00} & \textbf{0.00}  & 9540.92 & \textbf{100.00} & 84.25 & 15.75 & 15182.26 \\
& RDrop   & 99.50  & 69.00 & 30.65 & 20132.45 & \textbf{100.00} & 99.50 & 0.50 & 9516.64 & 99.75  & 84.25 & 15.54 & 14824.54 \\
& RanMASK & \textbf{100.00} & 68.50 & 31.50 & 21052.49 & \textbf{100.00} & 98.00 & 2.00 & \textbf{9748.31} & \textbf{100.00} & 83.25 & 16.75 & \textbf{15400.40} \\
& RMLM           & \textbf{100.00} & 71.50 & 28.50  & 20949.12 & \textbf{100.00} & 98.00  & 2.00  & 9373.92 & \textbf{100.00} & 84.75 & 15.25 & 15161.52 \\
\rowcolor{mygray}
& SCRN           & \textbf{100.00} & \textbf{82.50} & \textbf{17.50}  & \textbf{21122.83} & \textbf{100.00} & \textbf{100.00} & \textbf{0.00}  & 9540.20 & \textbf{100.00} & \textbf{91.25} & \textbf{8.75}  & 15331.52 \\
\bottomrule
\end{tabular}
}
\caption{Results of \textbf{in-domain} AIGT detection \textbf{under different adversarial attack methods}.}
\label{tab:in-domain-with-attack}
\end{table*}

\begin{table}[!h]
\centering
\scalebox{0.7}{
\begin{tabular}{lccccc}
\toprule
& P.(AI) & R.(AI) & P.(H.) & R.(H.) & F.(Overall) \\ 
\midrule
Log-Likelihood   & 95.88  & 76.99  & 80.78  & 96.67  & 86.71       \\
Log-Rank         & 94.86  & 72.21  & 77.57  & 96.08  & 83.92       \\
Entropy          & 97.56  & 48.84  & 65.88  & 98.78  & 72.06       \\
GLTR             & 87.61  & 71.85  & 76.14  & 89.84  & 80.69       \\
SeqXGPT          & 95.03  & 88.86  & 89.54  & 95.35  & 92.10       \\
Bert             & \textbf{99.54}  & 80.05  & 83.32  & \textbf{99.63}  & 89.74       \\
Roberta          & 82.77  & 87.03  & 86.32  & 86.32  & 84.45       \\
Deberta          & 97.99  & 89.35  & 90.21  & 98.16  & 93.75       \\
ChatGPT-Detector & 97.94  & \textbf{92.90}  & \textbf{93.25}  & 98.04  & \textbf{95.47}      \\
Flooding         & 84.19  & 88.00  & 87.44  & 83.48  & 85.73       \\
RDrop            & 90.43  & 87.88  & 88.21  & 90.70  & 89.29       \\
RanMASK          & 86.54  & 85.80  & 85.92  & 86.66  & 86.23       \\
RMLM             & 97.61  & 80.05  & 83.09  & 98.04  & 88.96       \\
\rowcolor{mygray} SCRN & 98.71  & 83.97  & 86.05  & 98.90  & 91.38       \\ 
\bottomrule
\end{tabular}
}
\caption{Results of \textbf{cross-domain} AIGT detection on \texttt{SeqXGPT-Bench} dataset \textbf{without attack}.}
\label{tab:cross-domain-without-attack}
\end{table}

\begin{table}[!h]
\centering
\scalebox{0.7}{
\begin{tabular}{lccccc}
\toprule
                 & P.(AI) & R.(AI) & P.(H.) & R.(H.) & F.(Overall) \\ 
\midrule
Log-Likelihood   & 92.60  & 75.03  & 79.01  & 94.00  & 84.38       \\
Log-Rank         & 95.58  & 63.43  & 72.64  & 97.07  & 79.68       \\
Entropy          & 72.45  & 75.03  & 74.11  & 71.47  & 73.24       \\
GLTR             & \textbf{96.36}  & 53.77  & 67.94  & \textbf{97.97}  & 74.63       \\
SeqXGPT          & 86.32  & 84.37  & 84.71  & 86.63  & \textbf{85.50}       \\
Bert             & 89.82  & 59.10  & 69.52  & 93.30  & 75.48       \\
Roberta          & 83.07  & 83.43  & 83.36  & 83.00  & 83.22       \\
Deberta          & 79.51  & 90.67  & 89.14  & 76.63  & 83.57       \\
ChatGPT-Detector & 86.46  & 69.40  & 69.08  & 90.70  & 74.42       \\
Flooding         & 83.75  & 87.60  & 87.00  & 83.00  & 85.29       \\
RDrop            & 77.05  & 94.43  & \textbf{92.81}  & 71.87  & 82.93       \\
RanMASK          & 89.84  & 68.10  & 74.32  & 92.30  & 79.90       \\
RMLM             & 89.90  & 57.83  & 68.92  & 93.50  & 74.87       \\
\rowcolor{mygray} 
SCRN & 73.61 & \textbf{94.67} & 92.53 & 66.07 & 79.96 \\ 
\bottomrule
\end{tabular}
}
\caption{Results of \textbf{cross-genre} AIGT detection on \texttt{SeqXGPT-Bench} dataset \textbf{without attack}.}
\label{tab:cross-genre-without-attack}
\end{table}

In the in-domain scenario, where the test data comes from the same domain as the training data (\texttt{HC3}), our SCRN model demonstrates superior robustness against various types of adversarial perturbations from three attack methods. As illustrated in Table \ref{tab:in-domain-with-attack}, SCRN achieves a notable improvement in Accuracy Under Attack (AUA), with a 6.5-11.25 absolute increase compared to the best baseline detector. The AUA values for SCRN remain high, exceeding 91.25 under all three attack methods, only marginally lower than its original accuracy without any attacks. Table \ref{tab:in-domain-with-attack} also reveals that \emph{evasion} attacks (tricking the model into classifying AI-generated answers as human, denoted by AI$\rightarrow$Human) are easier than \emph{obfuscation} attacks (tricking the model into classifying human-generated answers as AI, denoted by Human$\rightarrow$AI) on \texttt{HC3}. This is likely because human-generated answers are diverse \citep{ma2023ai}, making it challenging to perturb them all to resemble AI-generated responses.

In the absence of adversarial perturbation, all detectors except RanMASK demonstrate high performance, as illustrated in Table \ref{tab:in-domain-without-attack}. RanMASK's lower accuracy can be attributed to its masking of 30\% of the text during both training and inference, resulting in significant information loss.

\begin{table*}[!h]
\centering
\scalebox{0.7}{
\begin{tabular}{ll|cccc|cccc|cccc}
\toprule
 &
   &
  \multicolumn{4}{c|}{AI $\rightarrow$ Human} &
  \multicolumn{4}{c|}{Human $\rightarrow$ AI} &
  \multicolumn{4}{c}{Overall} \\
& Methods & OA $\uparrow$ & AUA $\uparrow$ & ASR $\downarrow$ & ANQ $\uparrow$ & OA $\uparrow$ & AUA $\uparrow$ & ASR $\downarrow$ & ANQ $\uparrow$ & OA $\uparrow$ & AUA $\uparrow$ & ASR $\downarrow$ & ANQ $\uparrow$ \\
\midrule
\multirow{13}{*}{\rotatebox{90}{PWWS}} &
  Log-Likelihood &
  67.00 &
  0.00 &
  100.00 &
  381.32 &
  97.00 &
  97.00 &
  0.00 &
  89.96 &
  82.00 &
  48.50 &
  40.85 &
  235.64 \\
 & Log-Rank  & 72.00 & 0.00  & 100.00 & 374.13  & 95.50 & 95.00 & 0.52  & 89.63  & 83.75 & 47.50 & 43.28 & 231.88  \\
 & Entropy   & 43.50 & 0.00  & 100.00 & 433.91  & 98.00 & 80.00 & 18.37 & 83.96  & 70.75 & 40.00 & 43.46 & 258.94  \\
 & GLTR      & 66.50 & 0.00  & 100.00 & 376.47  & 88.50 & 56.50 & 36.16 & 79.80  & 77.50 & 28.25 & 63.55 & 228.14  \\
 & SeqXGPT   & 93.00 & 4.00  & 95.70  & 397.55  & 98.50 & 94.00 & 4.56  & 97.64  & 95.75 & 49.00 & 48.83 & 247.60  \\
 & BERT      & 80.00 & 0.00  & 100.00 & 384.23  & 99.50 & 99.50 & 0.00  & 89.24  & 89.75 & 49.75 & 44.57 & 236.74  \\
 & RoBERTa   & 90.50 & 6.50  & 92.82  & 410.38  & 75.50 & 74.00 & 1.99  & \textbf{99.81}  & 83.00 & 40.25 & 51.51 & 255.10  \\
 & DeBERTa   & 91.50 & 1.00  & 98.91  & 381.80  & 98.00 & 97.50 & 0.51  & 88.78  & 94.75 & 49.25 & 48.02 & 235.29  \\
 & ChatGPT-Detector & \textbf{96.00} & 1.00  & 98.96  & 364.00  & 98.50 & 88.50 & 10.15 & 85.93  & \textbf{97.25} & 44.75 & 53.98 & 224.96  \\
 & Flooding  & 90.00 & 0.00 & 100.00 & 387.93 & 78.00 & 74.50 & 4.49 & 101.59 & 84.00 & 37.25 & 55.65 & 244.76  \\
 & RDrop & 89.50 & 6.00 & 93.30 & 429.06 & 89.00 & 73.50 & 17.42 & 91.10 & 89.25 & 39.75 & 55.46 & 260.08  \\
 & RanMASK   & 89.00 & 1.00  & 98.88  & 406.12  & 81.00 & 78.00 & 3.70  & 98.26  & 85.00 & 39.50 & 53.53 & 252.19  \\
 & RMLM      & 81.50 & 5.50  & 93.25  & 426.98  & 98.00 & 98.00 & 0.00  & 91.56  & 89.75 & 51.75 & 42.34 & 259.27  \\
 \rowcolor{mygray}
 & SCRN      & 86.50 & \textbf{40.50} & \textbf{53.18}  & \textbf{551.20}  & \textbf{99.50} & \textbf{99.50} & \textbf{0.00}  & 89.24  & 93.00 & \textbf{70.00} & \textbf{24.73} & \textbf{320.22}  \\
 \bottomrule
\end{tabular}
}
\caption{Results of \textbf{cross-domain} AIGT detection \textbf{under PWWS attack}. More results refer to Appendix \ref{sec:Details of Cross-domain AIGT Detection under Adversarial Perturbations}.}
\label{tab:cross-domain-results}
\end{table*}

\begin{table*}[!h]
\centering
\scalebox{0.7}{
\begin{tabular}{ll|cccc|cccc|cccc}
\toprule
 &
   &
  \multicolumn{4}{c|}{AI $\rightarrow$ Human} &
  \multicolumn{4}{c|}{Human $\rightarrow$ AI} &
  \multicolumn{4}{c}{Overall} \\
& Methods & OA $\uparrow$ & AUA $\uparrow$ & ASR $\downarrow$ & ANQ $\uparrow$ & OA $\uparrow$ & AUA $\uparrow$ & ASR $\downarrow$ & ANQ $\uparrow$ & OA $\uparrow$ & AUA $\uparrow$ & ASR $\downarrow$ & ANQ $\uparrow$ \\
\midrule
\multirow{13}{*}{\rotatebox{90}{PWWS}} 
 & Log-Likelihood & 62.00 & 0.00 & 100.00 & 2700.46 & \textbf{97.50} & \textbf{96.50} & \textbf{1.03} & \textbf{6077.26} & 79.75 & 48.25 & 39.50 & 4388.86 \\
 & Log-Rank  & 64.50 & 0.00  & 100.00 & 2734.98  & 97.50 & 95.50 & 2.05  & 6054.48  & 81.00 & 47.75 & 41.05 & 4394.73  \\
 & Entropy   & 77.50 & 0.00  & 100.00 & 2783.14  & 74.00 & 34.00 & 54.05 & 5352.47  & 75.75 & 17.00 & 77.56 & 4067.80  \\
 & GLTR      & 50.50 & 0.00  & 100.00 & 2696.80  & 97.50 & 67.50 & 30.77 & 5476.04  & 74.00 & 33.75 & 54.39 & 4086.42  \\
 & SeqXGPT   & 85.50 & 0.00  & 100.00 & 2712.97  & 88.00 & 65.50 & 25.57 & 5776.35  & \textbf{86.75} & 32.75 & 62.25 & 4244.66  \\
 & BERT      & 57.00 & 0.00  & 100.00 & 2692.61  & 95.50 & 75.00 & 21.47 & 5619.45  & 76.25 & 37.50 & 50.82 & 4156.03  \\
 & RoBERTa   & 82.00 & 0.00  & 100.00 & 2655.43  & 83.00 & 59.00 & 28.92 & 5522.05  & 82.50 & 29.50 & 64.24 & 4088.74  \\
 & DeBERTa   & 90.00 & 0.00  & 100.00 & 2763.66  & 77.50 & 53.50 & 30.97 & 5329.64  & 83.75 & 26.75 & 68.06 & 4046.65  \\
 & ChatGPT-Detector & 58.50 & 0.00  & 100.00 & 2606.75  & 93.00 & 73.00 & 21.51 & 5827.88  & 75.75 & 36.50 & 51.82 & 4217.32  \\
 & Flooding  & 87.50 & 0.00 & 100.00 & 2733.18 & 82.50 & 58.00 & 29.70 & 5447.84 & 85.00 & 29.00 & 65.88 & 4090.51  \\
 & RDrop & 95.00 & 10.00 & 89.47 & 3155.59 & 73.00 & 65.00 & 10.96 & 5973.84 & 84.00 & 37.50 & 55.36 & 4564.72 \\
 & RanMASK   & 67.00 & 2.00 & 97.01 & 2667.19 & 87.00 & 75.00 & 13.79 & 5433.82 & 77.00 & 38.50 & 50.00 & 4050.50  \\
 & RMLM      & 58.50 & 9.50  & 83.76  & 3397.99  & 92.00 & 72.50 & 21.20 & 5440.61  & 75.25 & 41.00 & 45.51 & 4419.30  \\
 \rowcolor{mygray}
 & SCRN      & \textbf{94.50} & \textbf{71.00} & \textbf{24.87}  & \textbf{4419.16}  & 70.50 & 54.50 & 22.70 & 5725.79  & 82.50 & \textbf{62.75} & \textbf{23.94} & \textbf{5072.48}  \\
 \bottomrule
\end{tabular}
}
\caption{Results of \textbf{cross-genre} AIGT detection \textbf{under PWWS attack}. More results refer to Appendix \ref{sec:Details of Cross-genre AIGT Detection under Adversarial Perturbations}.}
\label{tab:cross-genre-results}
\end{table*}

\begin{table*}[!h]
\centering
\scalebox{0.7}{
\begin{tabular}{ll|cccc|cccc|cccc}
\toprule
 &
   &
  \multicolumn{4}{c|}{AI $\rightarrow$ Human} &
  \multicolumn{4}{c|}{Human $\rightarrow$ AI} &
  \multicolumn{4}{c}{Overall} \\
& Methods & OA $\uparrow$ & AUA $\uparrow$ & ASR $\downarrow$ & ANQ $\uparrow$ & OA $\uparrow$ & AUA $\uparrow$ & ASR $\downarrow$ & ANQ $\uparrow$ & OA $\uparrow$ & AUA $\uparrow$ & ASR $\downarrow$ & ANQ $\uparrow$ \\
\midrule
\multirow{13}{*}{\rotatebox{90}{PWWS}}
 & Log-Likelihood & 72.00 & 0.50  & 99.31  & 1281.86  & 62.00 & 53.50 & 13.71 & 1667.91  & 67.00 & 27.00 & 59.70 & 1474.88  \\
 & Log-Rank       & 73.50 & 0.50  & 99.32  & 1286.24  & 62.50 & 56.00 & 10.40 & 1697.20  & 68.00 & 28.25 & 58.46 & 1491.72  \\
 & Entropy        & 63.00 & 0.00  & 100.00 & 1239.29  & 55.50 & 27.50 & 50.45 & 1396.39  & 59.25 & 13.75 & 76.79 & 1317.84  \\
 & GLTR           & 76.50 & 0.00  & 100.00 & 1260.99  & 67.50 & 19.00 & 71.85 & 1285.64  & 72.00 & 9.50  & 86.81 & 1273.32  \\
 & SeqXGPT        & \textbf{96.50} & 65.00 & 32.64  & 1867.81  & 96.00 & 70.00 & 27.08 & 1893.98  & \textbf{96.25} & 67.50 & 29.87 & 1880.90  \\
 & BERT           & 90.50 & 1.00  & 98.90  & 1204.52  & 90.00 & 59.00 & 34.44 & 1815.44  & 90.25 & 30.00 & 66.76 & 1509.98  \\
 & RoBERTa        & 95.50 & 64.50 & 32.46  & 1840.19  & 93.00 & 62.50 & 32.80 & 1729.72  & 94.25 & 63.50 & 32.63 & 1784.96  \\
 & DeBERTa        & 95.50 & 54.50 & 42.93  & 1764.94  & 96.00 & 80.00 & 16.67 & 1940.47  & 95.75 & 67.25 & 29.77 & 1852.70  \\
 & Flooding       & 96.00 & 60.50 & 36.98  & 1800.01  & 95.50 & 53.00 & 44.50 & 1610.45  & 95.75 & 56.75 & 40.73 & 1705.23  \\
 & RDrop & \textbf{96.50} & 69.00 & 28.50 & 1819.95 & 95.00 & 70.00 & 26.32 & 1815.62 & 95.75 & 69.50 & 27.42 & 1817.78 \\
 & RanMASK        & 94.00 & 60.00 & 36.17  & 1784.11  & 86.00 & 71.00 & 17.44 & 1715.72  & 90.00 & 65.50 & 27.22 & 1749.92  \\
 & RMLM           & 91.00 & 69.00 & 24.18  & 1879.96  & 91.50 & 78.00 & 14.75 & 1986.50  & 91.25 & 73.50 & 19.45 & 1933.23  \\
 \rowcolor{mygray}
 & SCRN           & 95.00 & \textbf{87.00} & \textbf{8.42}   & \textbf{1986.98}  & \textbf{96.00} & \textbf{91.50} & \textbf{4.69}  & \textbf{2099.91}  & 95.50 & \textbf{89.25} & \textbf{6.54}  & \textbf{2043.44}  \\
\bottomrule
\end{tabular}
}
\caption{Results of \textbf{mixed-source} AIGT detection \textbf{under PWWS attack}. The AI-generated texts are from five sources: GPT-2, GPT-Neo, GPT-J, LLaMa, and GPT-3. More results refer to Appendix \ref{sec:Details of Mixed-source AIGT Detection under Adversarial Perturbations}.}
\label{tab:mixed-source-with-attack}
\end{table*}

\subsection{Cross-domain Robustness} \label{sec:Cross-domain Robustness}

In the cross-domain setting, where the test data (\texttt{TruthfulQA}) and training data (\texttt{HC3}) come from different domains but share the same genre (question answering), SCRN significantly outperforms other baseline models. As Table \ref{tab:cross-domain-without-attack} and Table \ref{tab:cross-domain-results} demonstrates, SCRN achieves a notable improvement, with at least a +18.25 increase in absolute Accuracy Under Attack (AUA). Although absolute accuracy tends to be lower in cross-domain scenarios, SCRN's margin of lead is even more pronounced.

\subsection{Cross-genre Robustness} \label{sec:Cross-genre Robustness}

In the cross-genre setting, the test data (\texttt{Ghostbuster}) is from a different genre than the training data (\texttt{HC3}). Specifically, \texttt{Ghostbuster} comprises news articles, essays, and creative writings, while \texttt{HC3} is a question-answering dataset. As shown in Table \ref{tab:cross-genre-without-attack} and Table \ref{tab:cross-genre-results}, SCRN outperforms all baseline models in overall Accuracy Under Attack (AUA) and Attack Success Rate (ASR).

We observe that SCRN's AUA is lower than some baselines under obfuscation attacks (Human$\rightarrow$AI). However, SCRN is substantially more robust under evasion attacks (AI$\rightarrow$Human), achieving 71\% AUA, while baseline models' scores are near zero. This indicates that in the cross-genre setting, minor perturbations on AI-generated content can easily lead to a ``Human'' prediction by baseline models, as they are not trained on AI-generated content like news articles, essays, and creative writings. As a by-product, obfuscation attack against these models become very hard. Practically speaking, defending against evasion attacks is more important. SCRN demonstrates balanced robustness, underscoring its generalizability.

It is also noteworthy that RMLM, a model obtained through adversarial training, shows good robustness in the in-domain setting but fails in cross-domain and cross-genre settings. This suggests that merely augmenting the training set with adversarial data is insufficient to enhance the model's robustness against out-of-distribution samples \citep{wang2022improving}.

\subsection{Mixed-source Robustness} \label{sec:Mixed-source Robustness}

Tables \ref{tab:mixed-source-with-attack} and \ref{tab:mixed-source-without-attack} present the results on the \texttt{SeqXGPT-Bench} dataset, which comprises AI-generated content from various LLMs in mixed-source scenarios. Notably, SCRN shows excellent performance both with and without attacks.

Specifically, model-based detectors like SCRN significantly outperform metric-based detectors in the absence of an attack. However, the accuracy of all models decreases with the PWWS attack. Compared to the best-performing baseline, RMLM, our SCRN achieves a notable improvement of 15.75 in Accuracy Under Attack (AUA), consistent with other settings.

\begin{table}[!h]
\centering
\scalebox{0.7}{
\begin{tabular}{lccccc}
\toprule
               & P.(AI) & R.(AI) & P.(H.) & R.(H.) & F.(Overall) \\ \midrule
Log-Likelihood & 64.89  & 69.00  & 66.90  & 62.67  & 65.80       \\
Log-Rank       & 65.69  & 70.83  & 68.35  & 63.00  & 66.87       \\
Entropy        & 57.26  & 59.17  & 57.76  & 55.83  & 57.49       \\
GLTR           & 69.01  & 73.50  & 71.66  & 67.00  & 70.22       \\
SeqXGPT        & 95.03  & \textbf{97.39}  & \textbf{97.32}  & 94.93  & \textbf{96.15}       \\
Bert           & 86.76  & 90.67  & 90.23  & 86.17  & 88.50       \\
Roberta        & 94.77  & 96.66  & 96.60  & 94.67  & 95.67       \\
Deberta        & \textbf{95.99}  & 95.83  & 95.84  & \textbf{96.00}  & 95.92       \\
Flooding       & 94.91  & 96.33  & 96.28  & 94.83  & 95.58       \\
RDrop          & 94.63  & 97.00  & 96.92  & 94.50  & 95.75       \\
RanMASK        & 78.11  & 95.17  & 93.82  & 73.33  & 84.06       \\
RMLM           & 85.34  & 92.17  & 91.49  & 84.17  & 88.15       \\
\rowcolor{mygray}
SCRN    & 95.69  & 96.17  & 96.15  & 95.67  & 95.92       \\ \bottomrule
\end{tabular}
}
\caption{Results of \textbf{mixed-source} AIGT detection on \texttt{SeqXGPT-Bench} dataset \textbf{without attack}. The AI-generated texts are from five sources: GPT-2, GPT-Neo, GPT-J, LLaMa, and GPT-3.}
\label{tab:mixed-source-without-attack}
\end{table}

\subsection{Ablation Study} \label{sec:Ablation Study}

In this section, we perform an ablation study to evaluate the effectiveness of our key design choices. To assess the impact of siamese calibration, we train the SCRN model without it, denoted as SCRN-SC. Table \ref{tab:ablation-study} shows that the detector's robustness significantly worsens across all four AIGT scenarios. Notably, the accuracy under attack (AUA) of SCRN-SC is lower than the baseline RoBERTa detector (SCRN-SC-R) in the in-domain setting. This decline may be attributed to SCRN-SC's inclusion of random noise during inference, which, without siamese calibration, leads to unstable predictions under adversarial attacks. These findings underscore the importance of siamese calibration.

To examine the role of the reconstruction network, we replace it with a simple dropout layer, resulting in the SCRN-R model. As depicted in Table \ref{tab:ablation-study}, SCRN-R experiences a decrease in Accuracy Under Attack ranging from 18.0 to 33.25 across the four scenarios. This decline occurs because the dropout layer merely omits text information without simulating adversarial perturbations, which involve more complex editing actions such as substitution, insertion, and deletion.

Regarding the noise of the reconstruction network, completely removing the noise $\epsilon$ (SCRN-$\epsilon$) leads to a significant decrease in the detector's robustness, particularly in cross-domain scenarios of 29.5 AUA drop. Furthermore, when we dropped the regularization loss $\mathcal{L}_{reg}$, although the noise still existed, it degenerated during the optimization of $\mathcal{L}_{mse}$. The AUA scores of SCRN-$\mathcal{L}_{reg}$
demonstrated that maintaining a degenerated noise showed better robustness compared to completely removing the noise (SCRN-$\epsilon$). However, SCRN-$\mathcal{L}_{reg}$ performed significantly worse than SCRN, which highlights the necessity of the regularization loss $\mathcal{L}_{reg}$ in preserving the noise's effectiveness and maintaining the desired robustness.

\begin{table}[!h]
\centering
\scalebox{0.7}{
\begin{tabular}{ll|cccc}
\toprule
                              &                      & OA $\uparrow$ & AUA $\uparrow$ & ASR $\downarrow$ & ANQ $\uparrow$ \\ \hline
\multirow{6}{*}{In-domain}    & SCRN                 & 100.0         & 97.25          & 2.75             & 1445.28        \\
                              & -SC                  & 99.25         & 50.00          & 49.62            & 1067.74        \\
                              & -R                   & 99.50         & 79.25          & 20.35            & 1373.16        \\
                              & -$\epsilon$          & 100.00	       & 92.00	        & 8.00       	   & 1432.34        \\
                              & -$\mathcal{L}_{reg}$ & 100.00	       & 96.75          & 3.25             & 1444.54        \\
                              & -SC-R                & 100.0         & 69.00          & 31.00            & 1278.18        \\ \midrule
\multirow{6}{*}{Cross-domain} & SCRN                 & 93.00         & 70.00          & 24.73            & 320.22         \\
                              & -SC                  & 87.00         & 42.50          & 51.15            & 228.30         \\
                              & -R                   & 88.00         & 36.75          & 58.24            & 222.51         \\
                              & -$\epsilon$          & 82.50	       & 40.50	        & 50.91      	   & 280.56         \\
                              & -$\mathcal{L}_{reg}$ & 87.25         & 57.50          & 34.10            & 311.72         \\
                              & -SC-R                & 83.00         & 40.25          & 51.51            & 255.10         \\ \midrule
\multirow{6}{*}{Cross-genre}  & SCRN                 & 82.50         & 62.75          & 23.94            & 5072.48        \\
                              & -SC                  & 86.25         & 41.75          & 51.59            & 4318.61        \\
                              & -R                   & 84.50         & 35.00          & 58.58            & 4359.77        \\
                              & -$\epsilon$          & 83.00	       & 43.50	        & 47.59      	   & 4782.50        \\
                              & -$\mathcal{L}_{reg}$ & 79.50         & 59.50          & 25.16            & 4920.92        \\
                              & -SC-R                & 82.50         & 29.50          & 64.24            & 4088.74        \\ \midrule
\multirow{6}{*}{Mixed-source} & SCRN                 & 95.50         & 89.25          & 6.54             & 2043.44        \\
                              & -SC                  & 95.00         & 87.50          & 7.89             & 2025.96        \\
                              & -R                   & 96.00         & 68.00          & 29.17            & 1797.31        \\
                              & -$\epsilon$          & 95.75	       & 77.25	        & 19.32      	   & 1932.08        \\
                              & -$\mathcal{L}_{reg}$ & 96.00         & 82.75          & 13.80            & 1970.64        \\
                              & -SC-R                & 94.25         & 63.50          & 32.63            & 1784.96        \\ \bottomrule
\end{tabular}
}
\caption{The \textbf{ablation study} on four AIGT scenarios \textbf{under PWWS attack}.}
\label{tab:ablation-study}
\end{table}

\subsection{More Analysis}
We also provide a threshold analysis in Appendix \ref{sec:Threshold Analysis}, a comparison of inference speeds in Appendix \ref{sec:Inference Speed Comparison}, an analysis of inference fluctuations in Appendix \ref{sec:Inference Fluctuation Analysis}, and case studies in Appendix \ref{sec:Case Study}.

\section{Conclusion}
While AI-generated text (AIGT) detection is promising for various applications, it struggles with the robustness of current methods against adversarial perturbations. To tackle this, we introduce the Siamese Calibrated Reconstruction Network (SCRN). SCRN uses a reconstruction network to model text perturbations and employs siamese calibrated training to improve inference robustness. Experiments demonstrate SCRN's effectiveness in defending against adversarial perturbations in four AIGT scenarios, highlighting its practical utility in addressing real-world AIGT detection challenges.

\section*{Limitations}
Although SCRN demonstrates excellent robust performance across all four scenarios, including in-domain, cross-domain, cross-genre, and mixed-source AIGT detections, it still has several limitations:

(1) We did not consider the text paraphrasing attack \citep{tulchinskii2024intrinsic, macko2024authorship} as a form of text perturbation. Our focus was primarily on adversarial perturbations with minor modifications, and we regarded paraphrased text as completely modified. Strictly, if a paraphrased AI-generated text becomes the same as an existing human-created text, it should be assigned a human-created label by the AIGT detector. Future work may conduct a further more detailed analysis of the paraphrased text.

(2) Our experiments mainly focused on English corpora, and while our proposed method is general, we did not explore its performance on multilingual corpora. We leave the detailed analysis of multilingual datasets in future work.

\section*{Acknowledgements}
The work is supported by the National Key R\&D Program of China (Nos. 2022YFA1005201, 2022YFA1005202, 2022YFA1005203) and the NSFC Major Research Plan - Interpretable and General Purpose Next-generation Artificial Intelligence (No. 92270205).

\bibliography{custom}

\appendix

\section{Appendix}

\subsection{Datasets} \label{sec:Datasets}

\textbf{\texttt{HC3}} \citep{guo2023close}: We conducted our in-domain experiments using the \texttt{HC3} dataset, which is a compilation of human and ChatGPT answers from QA answers across four fields, including media, wiki, medicine, and finance. The dataset contains both English and Chinese data, but for our experiments, we focused on the English corpus. This corpus comprises 26,903 human texts and 58,546 ChatGPT texts. Following the methodology described in \citep{guo2023close}, we randomly partitioned the dataset, allocating 90\% for training and 10\% for testing. To evaluate the robustness of the AIGT detectors against adversarial perturbation attacks, we randomly selected 200 human-created samples and 200 AI-generated samples from the test set.

\textbf{\texttt{TruthfulQA}} \citep{he2023mgtbench}: To evaluate the detector robustness in cross-domain scenarios, we utilized the \texttt{TruthfulQA} dataset. This dataset comprises human-created answers and AI-generated answers for 817 questions spanning 38 diverse fields, such as law, finance, health, and politics. It provides a suitable environment for assessing the cross-domain robustness of AI-generated text methods. In our experiments, we trained all the compared detectors on the \texttt{HC3} train set and tested them on 817 human-created text and 817 ChatGPT-generated text of the \texttt{TruthfulQA} dataset. To evaluate the robustness of the detectors, we randomly selected 200 human-created samples and 200 ChatGPT-generated samples from \texttt{TruthfulQA} as test samples.

\textbf{\texttt{Ghostbuster}} \citep{verma2023ghostbuster}: For evaluating the detector robustness in cross-genre scenarios, we employed the \texttt{Ghostbuster} dataset. This dataset consists of 3,000 parallel human-created articles and AI-generated articles from student essays, news articles, and creative writing. Similarly, we trained all the compared detectors on the \texttt{HC3} train set, which compiles QA answers. We then assessed the detectors on articles from the \texttt{Ghostbuster} dataset. The robustness of detectors against adversarial attacks is evaluated by randomly selecting 200 human-created articles and 200 ChatGPT-generated articles from \texttt{Ghostbuster} dataset.

\textbf{\texttt{SeqXGPT-Bench}} \citep{wang-etal-2023-seqxgpt}: To assess the detector robustness under mixed-source AI-generated texts, we utilized the \texttt{SeqXGPT-Bench} dataset. This dataset consists of parallel human-created articles and AI-generated articles from various sources, including GPT-2, GPT-Neo, GPT-J, LLaMa, and GPT-3. It encompasses different domains such as news, social media, the web, scientific articles, and technical documents. For our experiments, we incorporated all 6,000 human-created texts and randomly selected one parallel AI-generated text from each of the five different AI sources, resulting in 6,000 parallel AI-generated texts from diverse sources. We allocated 90\% of the data for training and reserved the remaining 10\% as the test set. Again, for the adversarial perturbation evaluation, we randomly chose 200 human-created samples and 200 AI-generated samples from the test set.

\subsection{Adversarial Perturbations} \label{sec:Adversarial Perturbations}

To evaluate the impact of adversarial perturbations on the AIGT detectors, we leverage three adversarial attack methods that encompass character-level and word-level substitution, deletion, and insertion.

\textbf{PWWS} \citep{ren-etal-2019-generating}, a widely utilized adversarial attack method, efficiently performs synonym substitution based on word saliency scores and maximum word-swap variance.

\textbf{Deep-Word-Bug} \citep{gao-etal-2018-deepwordbug} incorporates random word-level substitution, swapping, deletion, and insertion, mimicking real-world human activities.

\textbf{Pruthi} \citep{pruthi-etal-2019-combating} introduces adversarial perturbations by altering a small number of characters, resembling common typos. It encompasses character substitution, deletion, and insertion.

\subsection{Baselines} \label{sec:Baselines}

To evaluate the performance of our proposed model, we compared it against several baseline models that are commonly used in AIGT detection. These baseline models include:

\textbf{Log-Likelihood} \citep{solaiman2019release} employs the average token-level log probability generated by a language model as a feature. It trains a machine-learning classifier to determine whether a human or machine generates the text. A higher log probability indicates a higher likelihood that the text is AI-generated.

\textbf{Log-Rank} \citep{mitchell-etal-2023-detectgpt} utilizes the logarithm of the probability rank of each token generated by a decoder-only language model. A lower rank suggests a higher likelihood that the text is AI-generated.

\textbf{Entropy} \citep{gehrmann-etal-2019-gltr} computes the entropy value for each token based on the preceding text and then averages these scores to create the final feature.

\textbf{GLTR} \citep{gehrmann-etal-2019-gltr} constructs features based on the number of tokens in the top-10, top-100, top-1000, and top-1000+ ranks according to the token-level probability rank generated by the language models. The idea is that AI-generated tokens are more likely to belong to the head of the distribution during decoding.

\textbf{SeqXGPT} \cite{wang-etal-2023-seqxgpt} ensembles log probability scores from multiple language models and trains a CNN-based transformer model to detect the AI-generated sentences.

\textbf{Bert} \citep{devlin-etal-2019-bert} is a widely used language model that demonstrates significant advancements in various NLP tasks, including many classification tasks. 

\textbf{Roberta} \citep{liu2019roberta} and \textbf{Deberta} \citep{he2020deberta} exhibit improved generalization ability compared to Bert, benefiting from additional training enhancements. 

\textbf{ChatGPT-Detector} \citep{guo2023close} is a RoBERTa-based detector tuned on the \texttt{HC3} dataset, which achieved state-of-the-art performance on ChatGPT text detection.

Additionally, we compared our method with recent robust methods that enhance classification robustness in other NLP tasks:

\textbf{Flooding} \citep{pmlr-v119-ishida20a} alleviates the model's overfitting by introducing an additional threshold into the training loss. This prevents the loss from decreasing further when it is already lower than the threshold, resulting in improved robustness.

\textbf{RDrop} \citep{wu2021r} leverages dropout layers to generate similar text representations and then aligns the outputs to be the same. It reveals better generalization in several NLP tasks. In our experiments, we use a RoBERTa-base encoder as the base model.

\textbf{RanMASK} \citep{zeng-etal-2023-certified} utilizes ensemble inference on several randomly masked text sequences copied from the input text. This approach enhances the model's robustness, particularly when encountering input word changes. In our experiments, we follow \citep{zeng-etal-2023-certified} to use a RoBERTa-base encoder as the base model and set the mask percentage to 30\%. 

\textbf{RMLM} \citep{wang-etal-2023-rmlm} is an adversarial training method that adds extra adversarial samples to the training data. It aims to defend the adversarial text attack by corrupting the adversarial text and then correcting the abnormal contexts. We follow the settings in their paper to train a BERT-based AIGT detector in our experiments.

\subsection{Experiment Settings} \label{sec:Experiment Settings}

We employed the RoBERTa-base encoder as the foundation of our model. To ensure a fair comparison, we selected the base versions of the pre-trained Bert, RoBERTa, and DeBERTa encoders used in the models under comparison. The detector training was conducted on 8 * 32GB NVIDIA V100 GPUs within the same environment. Subsequently, we evaluated the detectors under an adversarial perturbation attack on a single 32GB NVIDIA V100 GPU within the same environment.

Regarding the hyperparameters, we did not tune any of them in our experiment. For model training, we used a linear decay schedule with an initial learning rate of 1e-4. Consistent with \citep{guo2023close}, we trained all the compared detectors for 2 epochs on both the HC3 dataset and the SeqXGPT-Bench dataset. More detailed hyperparameter values can be found in Table \ref{tab:hyperparameters-app}.

To implement the adversarial attack methods, we utilized the open-source Textattack package \citep{morris2020textattack}. We implemented the compared models using their officially released code. For the ChatGPT-Detector, we directly utilized the model weight they released \footnote{\url{https://huggingface.co/Hello-SimpleAI/chatgpt-detector-roberta}}.

\begin{table}[!h]
\centering
\scalebox{1.0}{
\begin{tabular}{lc}
\toprule
\textbf{Hyperparameters} & \textbf{Value} \\ \midrule
Batch Size               & 16             \\
Training Epochs          & 2              \\
Optimizer                & AdamW          \\
Learning Rate            & 1e-4           \\
$d$                      & 768            \\
$d^z$                    & 512             \\
$\alpha$                 & 2.0            \\
$\beta$                  & 0.5            \\ 
$\lambda_1$              & 0.5            \\
$\lambda_2$              & 0.01           \\
$\lambda_3$              & 0.5            \\ \bottomrule
\end{tabular}
}
\caption{Hyperparameters of our SCRN AIGT detector.}
\label{tab:hyperparameters-app}
\end{table}

\subsection{Threshold Analysis} \label{sec:Threshold Analysis}
In Section \ref{sec:In-domain Robustness} - \ref{sec:Mixed-source Robustness}, we set the binary classification threshold to 0.5 to demonstrate the performance of AIGT detectors in general. However, the consequences of misclassifying human-created and AI-generated text in real-world scenarios may differ. Mislabeling human-created text as AI-generated text can have severe implications, such as affecting education exams and causing harm to innocent individuals. On the other hand, misclassifying AI-generated content as human-created can lead to the dissemination of harmful misinformation, resulting in public confusion or even social unrest.

\begin{table*}[t]
\centering
\scalebox{0.7}{
\begin{tabular}{l|cccc|cccc|cccc}
\toprule
& \multicolumn{4}{c|}{AI $\rightarrow$ Human} & \multicolumn{4}{c|}{Human $\rightarrow$ AI} & \multicolumn{4}{c}{Overall}     \\
Methods & OA $\uparrow$ &  AUA $\uparrow$ &  ASR $\downarrow$ & ANQ $\uparrow$ &  OA $\uparrow$ &  AUA $\uparrow$ &  ASR $\downarrow$ & ANQ $\uparrow$ &  OA $\uparrow$ &  AUA $\uparrow$ &  ASR $\downarrow$ &  ANQ $\uparrow$ \\ 
\midrule
Log-Likelihood & 8.50     & 0.00     & 100.00    & 1150.41   & 98.50     & 98.50    & \textbf{0.00}     & \textbf{2116.83}   & 53.50 & 49.25 & 7.94  & 1633.62 \\
Log-Rank       & 10.00    & 0.00     & 100.00    & 1212.20   & 99.50     & \textbf{99.00}    & 0.50     & 2102.61   & 54.75 & 49.50 & 9.59  & 1657.41 \\
Entropy        & 1.50     & 0.00     & 100.00    & 836.67    & 98.50     & 95.50    & 3.05     & 2106.64   & 50.00 & 47.75 & 4.50  & 1471.66 \\
GLTR           & 17.50    & 0.00     & 100.00    & 1407.69   & 99.50     & 83.00    & 16.58    & 1985.72   & 58.50 & 41.50 & 29.06 & 1696.70 \\
SeqXGPT        & 91.50    & 59.00    & 35.52     & 1787.49   & \textbf{100.00}    & 89.50    & 10.50    & 2018.60   & 95.75 & 74.25 & 22.45 & 1903.04 \\
BERT           & 73.00    & 0.00     & 100.00    & 1145.68   & 99.50     & 98.00    & 1.51     & 2097.26   & 86.25 & 49.00 & 43.19 & 1621.47 \\
RoBERTa        & 91.00    & 62.00    & 31.87     & 1788.83   & 99.00     & 76.50    & 22.73    & 1879.16   & 95.00 & 69.25 & 27.11 & 1834.00 \\
DeBERTa        & 91.50    & 46.50    & 49.18     & 1734.34   & \textbf{100.00}    & 88.00    & 12.00    & 2020.35   & 95.75 & 67.25 & 29.77 & 1877.34 \\
Flooding       & \textbf{92.50}    & 57.00    & 38.38     & 1782.71   & 98.50     & 65.00    & 34.01    & 1770.68   & 95.50 & 61.00 & 36.13 & 1776.70 \\
RDrop          & \textbf{92.50}    & 68.00    & 26.49     & 1819.03   & 99.50     & 83.00    & 16.58    & 1914.17   & \textbf{96.00} & 75.50 & 21.35 & 1866.60 \\
RMLM           & 84.50    & 68.00    & 19.53     & 1811.02   & \textbf{100.00}    & 94.00    & 6.00     & 2024.73   & 92.25 & 81.00 & 12.20 & 1917.88 \\
\rowcolor{mygray}
SCRN           & 89.00    & \textbf{81.00}    & \textbf{8.99}      & \textbf{1972.56}   & \textbf{100.00}    & \textbf{99.00}    & 1.00     & 2101.37   & 94.50 & \textbf{90.00} & \textbf{4.76}  & \textbf{2036.96} \\ 
\bottomrule
\end{tabular}
}
\caption{AIGT detection results under fixed 1\% false positive rate (FPR) considering \textbf{AI-generated text as positive samples}. Detectors are trained on the \texttt{Seqxgpt-Bench} training set, tested on randomly selected 200 human-created text and 200 AI-generated text from the \texttt{Seqxgpt-Bench} test set, and subjected to attacks using the PWWS method on the \texttt{Seqxgpt-Bench} attack set. The threshold is chosen to maintain an FPR of 1\% across the entire test set. The best results are highlighted in bold.}
\label{tab:ai-pos-1-FPR}
\end{table*}

\begin{table*}[t]
\centering
\scalebox{0.7}{
\begin{tabular}{l|cccc|cccc|cccc}
\toprule
& \multicolumn{4}{c|}{AI $\rightarrow$ Human} & \multicolumn{4}{c|}{Human $\rightarrow$ AI} & \multicolumn{4}{c}{Overall}     \\
Methods & OA $\uparrow$ &  AUA $\uparrow$ &  ASR $\downarrow$ & ANQ $\uparrow$ &  OA $\uparrow$ &  AUA $\uparrow$ &  ASR $\downarrow$ & ANQ $\uparrow$ &  OA $\uparrow$ &  AUA $\uparrow$ &  ASR $\downarrow$ &  ANQ $\uparrow$ \\ 
\midrule
Log-Likelihood & 99.50     & 1.50     & 98.49    & 1521.43   & 0.50     & 0.00     & 100.00    & 504.00    & 50.00 & 0.75  & 98.50 & 1012.72 \\
Log-Rank       & 99.50     & 1.00     & 98.99    & 1517.97   & 2.00     & 0.50     & 75.00     & 470.75    & 50.75 & 0.75  & 98.52 & 994.36  \\
Entropy        & 98.50     & 0.50     & 99.49    & 1292.23   & 5.00     & 0.00     & 100.00    & 590.50    & 51.75 & 0.25  & 99.52 & 941.36  \\
GLTR           & 99.50     & 1.50     & 98.49    & 1532.50   & 1.00     & 0.00     & 100.00    & 362.00    & 50.25 & 0.75  & 98.51 & 947.25  \\
SeqXGPT        & 99.50     & 77.00    & 22.61    & 1934.58   & \textbf{88.50}    & 54.50    & 38.42     & 1636.77   & \textbf{94.00} & 65.75 & 30.05 & 1785.68 \\
BERT           & 99.00     & 11.00    & 88.89    & 1299.73   & 35.50    & 11.00    & 69.01     & 1981.18   & 67.25 & 11.00 & 83.64 & 1640.46 \\
RoBERTa        & 99.00     & 75.50    & 23.74    & 1913.65   & 82.00    & 48.50    & 40.85     & 1640.80   & 90.50 & 62.00 & 31.49 & 1777.22 \\
DeBERTa        & 99.00     & 71.00    & 28.28    & 1880.54   & \textbf{88.50}    & 56.50    & 36.16     & 1648.58   & 93.75 & 63.75 & 32.00 & 1764.56 \\
Flooding       & 99.50     & 69.50    & 30.15    & 1846.76   & 86.00    & 41.00    & 52.33     & 1449.56   & 92.75 & 55.25 & 40.43 & 1648.16 \\
RDrop          & \textbf{100.00}    & 74.50    & 25.50    & 1832.50   & 88.00    & 62.00    & 29.55     & 1742.41   & \textbf{94.00} & 68.25 & 27.39 & 1787.46 \\
RMLM           & 99.00     & 80.50    & 18.69    & 1965.81   & 81.00    & 59.00    & 27.16     & 1678.57   & 90.00 & 69.75 & 22.50 & 1822.19 \\
\rowcolor{mygray}
SCRN           & 98.50     & \textbf{95.00}    & \textbf{3.55}     & \textbf{2013.52}   & 86.50    & \textbf{72.00}    & \textbf{16.76}     & \textbf{1857.72}   & 92.50 & \textbf{83.50} & \textbf{9.73}  & \textbf{1935.62} \\
\bottomrule
\end{tabular}
}
\caption{AIGT detection results under fixed 1\% false positive rate (FPR) regarding \textbf{human-created text as positive samples}. Detectors are trained on the \texttt{Seqxgpt-Bench} training set, tested on randomly selected 200 human-created text and 200 AI-generated text from the \texttt{Seqxgpt-Bench} test set, and subjected to attacks using the PWWS method on the \texttt{Seqxgpt-Bench} attack set. The threshold is chosen to maintain an FPR of 1\% across the entire test set. The best results are highlighted in bold.}
\label{tab:human-pos-1-FPR}
\end{table*}

To evaluate the robustness of AIGT detectors under different misclassification cost variations, we conducted experiments following the 1\% false positive rate (FPR) setting proposed in \citep{hans2024spotting, krishna2024paraphrasing, soto2024few}. Specifically, we evaluated the detectors under a fixed 1\% FPR, where either AI-generated or human-created text is considered the positive sample. The training and testing were performed on \texttt{SeqXGPT-Bench}, which involves the generation of AI-generated text by mixed-source language models, better simulating the challenges encountered in real-world applications.

As shown in Table \ref{tab:ai-pos-1-FPR}, when considering AI-generated text as positive and using the threshold under the 1\% FPR setting, our SCRN detector achieves the highest AUA score, signifying superior robustness, while remaining competitive in terms of the OA score. Similarly, as illustrated in Table \ref{tab:human-pos-1-FPR}, when setting the threshold for 1\% FPR with human-created text as positive samples, SCRN consistently demonstrates the best robustness, as indicated by the AUA score, and maintains competitive accuracy without attack, as measured by the OA score.

\subsection{Inference Speed Comparison} \label{sec:Inference Speed Comparison}

\begin{table}[!h]
\centering
\scalebox{0.7}{
\begin{tabular}{ll|cccc}
\toprule
                              &         & OA $\uparrow$ & AUA $\uparrow$ & ASR $\downarrow$ & ANQ $\uparrow$ \\
\midrule
\multirow{4}{*}{In-domain}    & RoBERTa & \textbf{100.00}        & 74.67          & 25.33            & 5524.22        \\
                              & RanMASK & \textbf{100.00}        & 79.25          & 20.75            & 5709.73        \\
                              & RMLM    & \textbf{100.00}        & 84.58          & 15.42            & 5631.18        \\
                              \rowcolor{mygray}
                              & SCRN    & \textbf{100.00}        & \textbf{94.08}          & \textbf{5.92}             & \textbf{5716.17}        \\
\midrule
\multirow{4}{*}{Corss-domain} & RoBERTa & 83.00         & 42.58          & 48.70            & 590.99         \\
                              & RanMASK & 85.00         & 45.50          & 46.47            & 462.53         \\
                              & RMLM    & 89.75         & 51.58          & 42.53            & 535.79         \\
                              \rowcolor{mygray}
                              & SCRN    & \textbf{93.00}         & \textbf{64.50}          & \textbf{30.65}            & \textbf{638.54}         \\
\midrule
\multirow{4}{*}{Cross-genre}  & RoBERTa & \textbf{82.50}         & 31.00          & 62.42            & 17991.85       \\
                              & RanMASK & 77.00         & 30.25          & 60.71            & 15757.80       \\
                              & RMLM    & 75.25         & 40.08          & 46.73            & 20728.78       \\
                              \rowcolor{mygray}
                              & SCRN    & 82.25         & \textbf{55.50}          & \textbf{32.53}            & \textbf{23179.41}       \\
\midrule
\multirow{4}{*}{Mixed-source} & RoBERTa & 94.25         & 59.58          & 36.78            & 9644.89        \\
                              & RanMASK & 90.00         & 62.25          & 30.83            & 10127.65       \\
                              & RMLM    & 91.25         & 72.17          & 20.91            & 11765.38       \\
                              \rowcolor{mygray}
                              & SCRN    & \textbf{95.33}         & \textbf{85.33}          & \textbf{10.50}            & \textbf{13115.02}      \\
\bottomrule
\end{tabular}
}
\caption{The robustness comparison of RoBERTa, RanMASK, RMLM, and SCRN. Results are average scores under three types of adversarial perturbation attacks including PWWS, Deep-Word-Bug, and Pruthi.}
\label{tab:selected-model-comparison}
\end{table}

\begin{figure}[h]
\centering 
\includegraphics[width = 0.45\textwidth]{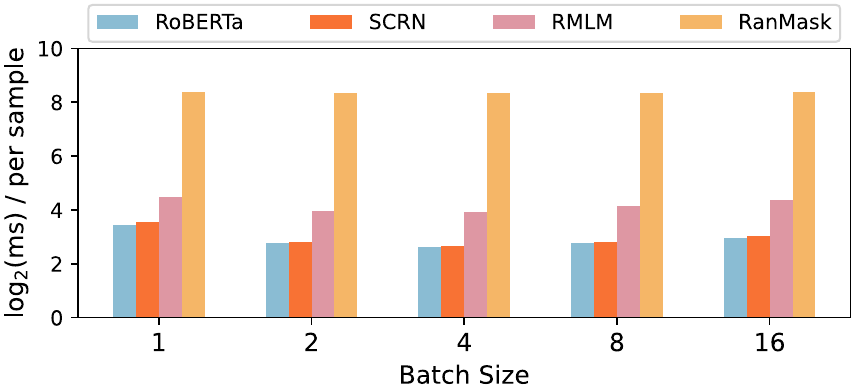} 
\caption{Inference time comparison of RoBERTa, RanMASK, RMLM, and SCRN on \texttt{HC3} test set. The experiments are conducted on a single 32GB NVIDIA-V100 GPU.} 
\label{fig:infer-speed} 
\end{figure}

Since the inference speed is an essential metric to evaluate the AIGT detector, we compared the inference speed of our proposed SCRN model with three other models: the RoBERTa backbone model, the previous state-of-the-art robust method RanMASK, and the adversarial training method RMLM. 

Our SCRN model demonstrates highly effective inference speed when compared to the other AIGT models. As presented in Figure \ref{fig:infer-speed}, in real-world inference service scenarios with a batch size of one, our model achieves an inference speed that is $1.91 \times$ faster than RMLM. This speed advantage is primarily because RMLM involves result selection from a twice-forward process. Additionally, our model achieves an inference speed that is $28.5 \times$ faster than RanMASK, as RanMASK requires ensembling hundreds of results. When compared to the RoBERTa backbone detector, our model maintains $92.7\%$ of RoBERTa's inference speed while surpassing up to 25.75 of accuracy under attack (AUA) achieved by RoBERTa as shown in Table \ref{tab:selected-model-comparison}.

For larger batch sizes, our SCRN model also demonstrates significantly faster inference speed compared to RMLM and RanMASK, with only a slight decrease in speed when compared to RoBERTa.

\subsection{Inference Fluctuation Analysis} \label{sec:Inference Fluctuation Analysis}

\begin{figure}[h]
\centering 
\includegraphics[width = 0.45\textwidth]{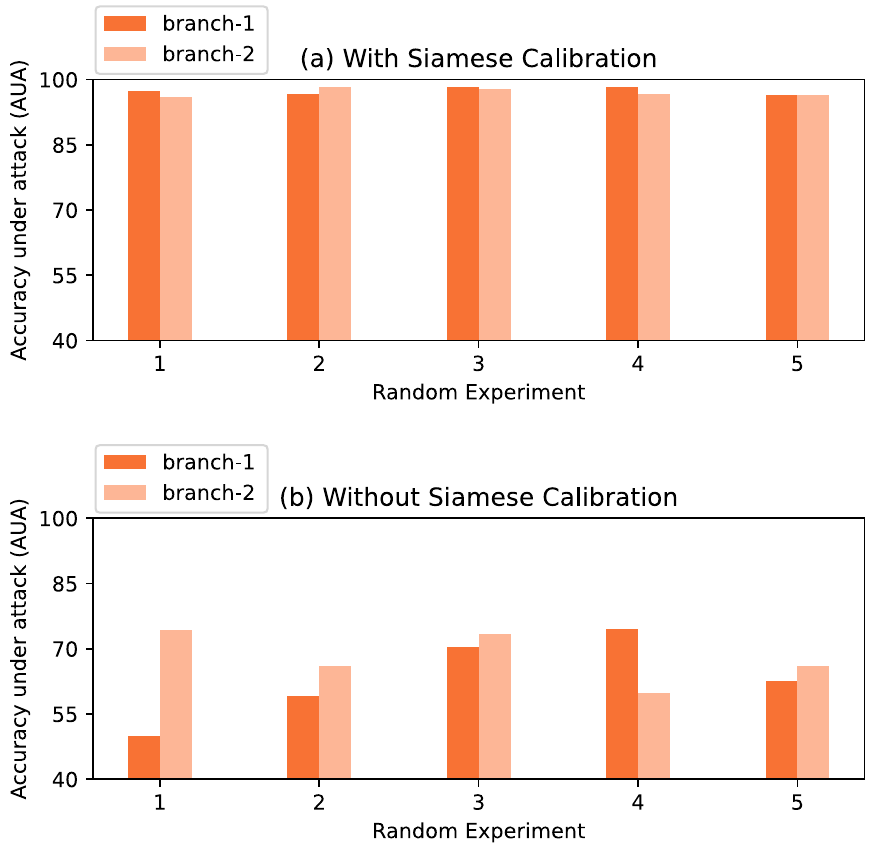} 
\caption{Inference fluctuation between two sub branches of SCRN
on \texttt{HC3} test set.} 
\label{fig:infer-fluctuation} 
\end{figure}

To assess the impact of inference fluctuation between two sub-model branches of SCRN, particularly due to randomness introduced by the reconstruction layer, we conducted a comparison of the detectors' accuracy under attack (AUA). Figure \ref{fig:infer-fluctuation} presents the AUA fluctuation on \texttt{HC3} dataset, considering five different random seeds. Notably, the AUA fluctuation remains below a 2.5 accuracy score across all scenarios. This observation indicates that the randomness introduced by the reconstruction process does not significantly affect the robustness of the SCRN detector. Thus, the gap between training and inference can be neglected.

This resilience can be attributed to the implementation of siamese calibration during the training of SCRN. The siamese calibration strategy aims to minimize the discrepancy in output distributions between the two branches of the model. Consequently, the classifier layer of SCRN is encouraged to prioritize high-level features over token-level features. This preference for high-level features enhances the model's robustness against token-level noise originating from the reconstruction process.

In contrast, when SCRN is trained without siamese calibration, the AUA fluctuation increases significantly, reaching up to a 14.75 accuracy score. This ablation experiment serves as further evidence of the effectiveness of siamese calibration in enhancing the robustness of the detector.

\subsection{Case Study} \label{sec:Case Study}

\begin{figure*}[t]
\centering 
\includegraphics[width = 0.95\textwidth]{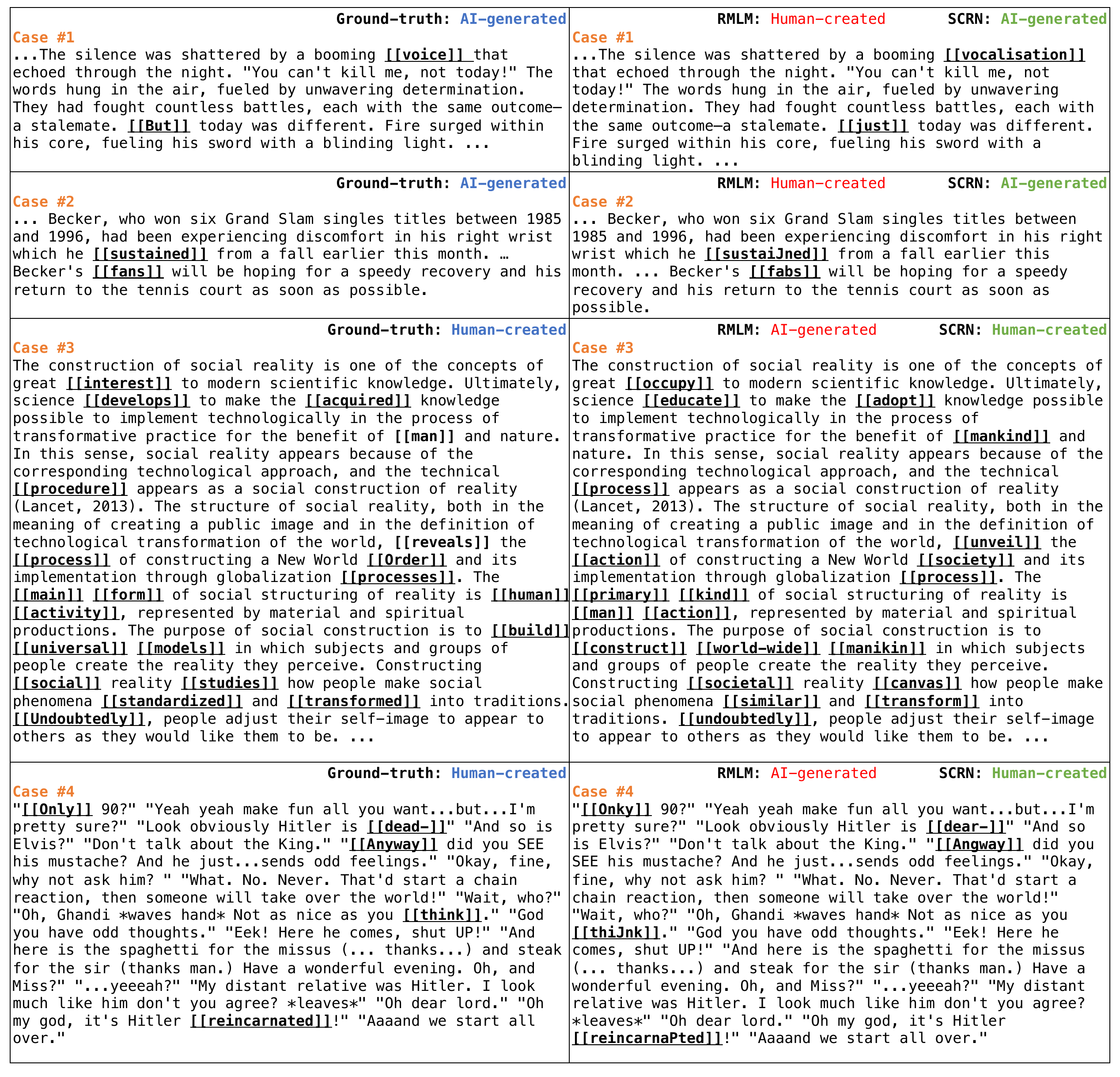} 
\caption{Cases from the \texttt{Ghostbuster} dataset are depicted in the figure. Case \#1 and \#2 represent AI-generated samples, whereas Case \#3 and \#4 are human-created samples. In cross-genre scenarios, RMLM fails to defend against adversarial text perturbations, whereas our SCRN demonstrates superior robustness. These cases highlight successful attacks on RMLM, while all adversarial attacks on these texts are unsuccessful against SCRN. Perturbed words or characters are \text{\underline{[[\textbf{highlighted}]]}}, while unchanged text is omitted for clarity.} 
\label{fig:case-study} 
\end{figure*}

As part of our research, we conducted a case study involving four cases from the \texttt{Ghostbuster} dataset, representing a cross-genre scenario. In this study, we compared our model with the previous state-of-the-art robust model, RMLM, which has been trained using adversarial techniques.

Figure \ref{fig:case-study} illustrates Case \#1 and Case \#2, which demonstrate word perturbations and character perturbations, respectively. While RMLM improves its robustness against adversarial perturbations by incorporating additional adversarial data during training, it fails to maintain robust detection in the cross-genre scenario. This observation highlights that simply augmenting the training set with adversarial data is not sufficient to effectively enhance the model's robustness against out-of-distribution samples \citep{wang2022improving}. In contrast, our proposed SCRN model does not rely on extra prior information from the training set. As a result, SCRN successfully defends against cross-genre attacks, achieving superior robustness performance, as demonstrated in Table \ref{tab:details-cross-genre-with-attack}.

Moving on to Case \#3 and Case \#4, these cases explore the effects of word perturbations and character perturbations in human-created text, respectively. In comparison to AI-generated text, human-created text poses a greater challenge for attacks. It requires a higher proportion of manipulated words or characters to deceive the AIGT detector successfully.

\subsection{Details of Cross-domain AIGT Detection under Adversarial Perturbations} \label{sec:Details of Cross-domain AIGT Detection under Adversarial Perturbations}

Table \ref{tab:details-cross-domain-with-attack} shows the robustness performance of all compared AIGT detectors on \texttt{TruthfulQA} dataset under three adversarial attack methods. 
The results consistently demonstrate the superior robustness of SCRN in cross-domain AIGT detection.

\begin{table*}[!h]
\centering
\scalebox{0.7}{
\begin{tabular}{ll|cccc|cccc|cccc}
\toprule
 &
   &
  \multicolumn{4}{c|}{AI $\rightarrow$ Human} &
  \multicolumn{4}{c|}{Human $\rightarrow$ AI} &
  \multicolumn{4}{c}{Overall} \\
& Methods & OA $\uparrow$ & AUA $\uparrow$ & ASR $\downarrow$ & ANQ $\uparrow$ & OA $\uparrow$ & AUA $\uparrow$ & ASR $\downarrow$ & ANQ $\uparrow$ & OA $\uparrow$ & AUA $\uparrow$ & ASR $\downarrow$ & ANQ $\uparrow$ \\
\midrule
\multirow{13}{*}{\rotatebox{90}{PWWS}} &
  Log-Likelihood &
  67.00 &
  0.00 &
  100.00 &
  381.32 &
  97.00 &
  97.00 &
  0.00 &
  89.96 &
  82.00 &
  48.50 &
  40.85 &
  235.64 \\
 & Log-Rank  & 72.00 & 0.00  & 100.00 & 374.13  & 95.50 & 95.00 & 0.52  & 89.63  & 83.75 & 47.50 & 43.28 & 231.88  \\
 & Entropy   & 43.50 & 0.00  & 100.00 & 433.91  & 98.00 & 80.00 & 18.37 & 83.96  & 70.75 & 40.00 & 43.46 & 258.94  \\
 & GLTR      & 66.50 & 0.00  & 100.00 & 376.47  & 88.50 & 56.50 & 36.16 & 79.80  & 77.50 & 28.25 & 63.55 & 228.14  \\
 & SeqXGPT   & 93.00 & 4.00  & 95.70  & 397.55  & 98.50 & 94.00 & 4.56  & 97.64  & 95.75 & 49.00 & 48.83 & 247.60  \\
 & BERT      & 80.00 & 0.00  & 100.00 & 384.23  & 99.50 & 99.50 & 0.00  & 89.24  & 89.75 & 49.75 & 44.57 & 236.74  \\
 & RoBERTa   & 90.50 & 6.50  & 92.82  & 410.38  & 75.50 & 74.00 & 1.99  & 99.81  & 83.00 & 40.25 & 51.51 & 255.10  \\
 & DeBERTa   & 91.50 & 1.00  & 98.91  & 381.80  & 98.00 & 97.50 & 0.51  & 88.78  & 94.75 & 49.25 & 48.02 & 235.29  \\
 & ChatGPT-Detector & 96.00 & 1.00  & 98.96  & 364.00  & 98.50 & 88.50 & 10.15 & 85.93  & 97.25 & 44.75 & 53.98 & 224.96  \\
 & Flooding  & 90.00 & 0.00 & 100.00 & 387.93 & 78.00 & 74.50 & 4.49 & 101.59 & 84.00 & 37.25 & 55.65 & 244.76  \\
 & RDrop & 89.50 & 6.00 & 93.30 & 429.06 & 89.00 & 73.50 & 17.42 & 91.10 & 89.25 & 39.75 & 55.46 & 260.08  \\
 & RanMASK   & 89.00 & 1.00  & 98.88  & 406.12  & 81.00 & 78.00 & 3.70  & 98.26  & 85.00 & 39.50 & 53.53 & 252.19  \\
 & RMLM      & 81.50 & 5.50  & 93.25  & 426.98  & 98.00 & 98.00 & 0.00  & 91.56  & 89.75 & 51.75 & 42.34 & 259.27  \\
 \rowcolor{mygray}
 & SCRN      & 86.50 & 40.50 & 53.18  & 551.20  & 99.50 & 99.50 & 0.00  & 89.24  & 93.00 & 70.00 & 24.73 & 320.22  \\
\midrule
\multirow{13}{*}{\rotatebox{90}{Deep-Word-Bug}} &
  Log-Likelihood &
  67.00 &
  0.00 &
  100.00 &
  47.54 &
  97.00 &
  97.00 &
  0.00 &
  27.45 &
  82.00 &
  48.50 &
  40.85 &
  37.49 \\
 & Log-Rank  & 72.00 & 0.00  & 100.00 & 47.07   & 95.50 & 95.50 & 0.00  & 27.39  & 83.75 & 47.75 & 42.99 & 37.23   \\
 & Entropy   & 43.50 & 0.00  & 100.00 & 54.54   & 98.00 & 95.00 & 3.06  & 26.86  & 70.75 & 47.50 & 32.86 & 40.70   \\
 & GLTR      & 66.50 & 0.00  & 100.00 & 48.53   & 88.50 & 84.00 & 5.08  & 25.86  & 77.50 & 42.00 & 45.81 & 37.20   \\
 & SeqXGPT   & 93.00 & 0.50  & 99.46  & 56.35   & 98.50 & 98.50 & 0.00  & 27.13  & 95.75 & 49.50 & 48.30 & 41.74   \\
 & BERT      & 80.00 & 0.00  & 100.00 & 52.19   & 99.50 & 99.50 & 0.00  & 27.51  & 89.75 & 49.75 & 44.57 & 39.85   \\
 & RoBERTa   & 90.50 & 4.50  & 95.03  & 67.50   & 75.50 & 75.00 & 0.66  & 30.64  & 83.00 & 39.75 & 52.11 & 49.07   \\
 & DeBERTa   & 91.50 & 2.00  & 97.81  & 53.75   & 98.00 & 98.00 & 0.00  & 27.33  & 94.75 & 50.00 & 47.23 & 40.54   \\
 & ChatGPT-Detector & 96.00 & 0.00  & 100.00 & 55.81   & 98.50 & 97.50 & 1.02  & 27.38  & 97.25 & 48.75 & 49.87 & 41.60   \\
 & Flooding  & 90.00 & 1.50  & 98.33  & 57.78   & 78.00 & 76.50 & 1.92  & 31.06  & 84.00 & 39.00 & 53.57 & 44.42   \\
 & RDrop & 89.50  & 4.00  & 95.53 & 73.26   & 89.00  & 88.00  & 1.12 & 28.09   & 89.25  & 46.00 & 48.46 & 50.68   \\
 & RanMASK   & 89.00 & 3.00 & 96.63 & 68.10 & 81.00 & 75.00 & 7.41 & 27.84 & 85.00 & 39.00 & 54.12 & 47.97   \\
 & RMLM      & 81.50 & 2.50  & 96.93  & 56.86   & 98.00 & 98.00 & 0.00  & 28.35  & 89.75 & 50.25 & 44.01 & 42.60   \\
 \rowcolor{mygray}
 & SCRN      & 86.50 & 17.00 & 80.35  & 107.34  & 99.50 & 99.50 & 0.00  & 27.50  & 93.00 & 58.25 & 37.37 & 67.42   \\
\midrule
\multirow{13}{*}{\rotatebox{90}{Pruthi}} &
  Log-Likelihood &
  67.00 &
  1.00 &
  98.51 &
  1849.10 &
  97.00 &
  97.00 &
  0.00 &
  151.68 &
  82.00 &
  49.00 &
  40.24 &
  1000.39 \\
 & Log-Rank  & 72.00 & 2.00  & 97.22  & 1938.40 & 95.50 & 95.50 & 0.00  & 151.69 & 83.75 & 48.75 & 41.79 & 1045.04 \\
 & Entropy   & 43.50 & 1.00  & 97.70  & 2329.26 & 98.00 & 93.50 & 4.59  & 150.90 & 70.75 & 47.25 & 33.22 & 1240.08 \\
 & GLTR      & 66.50 & 1.00  & 98.50  & 2025.11 & 88.50 & 82.00 & 7.34  & 147.90 & 77.50 & 41.50 & 46.45 & 1086.50 \\
 & SeqXGPT   & 93.00 & 3.50  & 96.24  & 2265.47 & 98.50 & 98.50 & 0.00  & 150.38 & 95.75 & 51.00 & 46.74 & 1207.92 \\
 & BERT      & 80.00 & 8.50  & 89.38  & 2733.19 & 99.50 & 99.50 & 0.00  & 151.35 & 89.75 & 54.00 & 39.83 & 1442.27 \\
 & RoBERTa   & 90.50 & 21.50 & 76.24  & 2767.34 & 75.50 & 74.00 & 1.99  & 170.28 & 83.00 & 47.75 & 42.47 & 1468.81 \\
 & DeBERTa   & 91.50 & 7.00  & 92.35  & 2387.38 & 98.00 & 97.00 & 1.02  & 150.17 & 94.75 & 52.00 & 45.12 & 1268.78 \\
 & ChatGPT-Detector & 96.00 & 17.50 & 81.77  & 2693.03 & 98.50 & 91.00 & 7.61  & 151.59 & 97.25 & 54.25 & 44.22 & 1422.31 \\
 & Flooding  & 90.00 & 25.00 & 72.22  & 2873.47 & 78.00 & 76.50 & 1.92  & 172.08 & 84.00 & 50.75 & 39.58 & 1522.78 \\
 & RDrop & 89.50  & 24.50 & 72.63 & 2813.49 & 89.00  & 87.00  & 2.25 & 155.13  & 89.25  & 55.75 & 37.54 & 1484.31 \\
 & RanMASK   & 89.00 & 36.00 & 59.55 & 2011.91 & 81.00 & 80.00 & 1.23 & 162.98 & 85.00 & 58.00 & 31.76 & 1087.44 \\
 & RMLM      & 81.50 & 7.50  & 90.80  & 2451.19 & 98.00 & 98.00 & 0.00  & 159.80 & 89.75 & 52.75 & 41.23 & 1305.50 \\
 \rowcolor{mygray}
 & SCRN      & 86.50 & 31.00 & 64.16  & 2904.60 & 99.50 & 99.50 & 0.00  & 151.35 & 93.00 & 65.25 & 29.84 & 1527.98 \\
\bottomrule
\end{tabular}
}
\caption{Results of cross-domain AIGT detection under different attack methods.}
\label{tab:details-cross-domain-with-attack}
\end{table*}

\subsection{Details of Cross-genre AIGT Detection under Adversarial Perturbations} \label{sec:Details of Cross-genre AIGT Detection under Adversarial Perturbations}

Table \ref{tab:details-cross-genre-with-attack} shows the robustness performance of all compared AIGT detectors on \texttt{Ghostbuster} dataset under three adversarial attack methods. The results consistently demonstrate the superior robustness of SCRN in cross-genre AIGT detection.

\begin{table*}[!h]
\centering
\scalebox{0.7}{
\begin{tabular}{ll|cccc|cccc|cccc}
\toprule
 &
   &
  \multicolumn{4}{c|}{AI $\rightarrow$ Human} &
  \multicolumn{4}{c|}{Human $\rightarrow$ AI} &
  \multicolumn{4}{c}{Overall} \\
& Methods & OA $\uparrow$ & AUA $\uparrow$ & ASR $\downarrow$ & ANQ $\uparrow$ & OA $\uparrow$ & AUA $\uparrow$ & ASR $\downarrow$ & ANQ $\uparrow$ & OA $\uparrow$ & AUA $\uparrow$ & ASR $\downarrow$ & ANQ $\uparrow$ \\
\midrule
\multirow{13}{*}{\rotatebox{90}{PWWS}} &
  Log-Likelihood &
  62.00 &
  0.00 &
  100.00 &
  2700.46 &
  97.50 &
  96.50 &
  1.03 &
  6077.26 &
  79.75 &
  48.25 &
  39.50 &
  4388.86 \\
 & Log-Rank  & 64.50 & 0.00  & 100.00 & 2734.98  & 97.50 & 95.50 & 2.05  & 6054.48  & 81.00 & 47.75 & 41.05 & 4394.73  \\
 & Entropy   & 77.50 & 0.00  & 100.00 & 2783.14  & 74.00 & 34.00 & 54.05 & 5352.47  & 75.75 & 17.00 & 77.56 & 4067.80  \\
 & GLTR      & 50.50 & 0.00  & 100.00 & 2696.80  & 97.50 & 67.50 & 30.77 & 5476.04  & 74.00 & 33.75 & 54.39 & 4086.42  \\
 & SeqXGPT   & 85.50 & 0.00  & 100.00 & 2712.97  & 88.00 & 65.50 & 25.57 & 5776.35  & 86.75 & 32.75 & 62.25 & 4244.66  \\
 & BERT      & 57.00 & 0.00  & 100.00 & 2692.61  & 95.50 & 75.00 & 21.47 & 5619.45  & 76.25 & 37.50 & 50.82 & 4156.03  \\
 & RoBERTa   & 82.00 & 0.00  & 100.00 & 2655.43  & 83.00 & 59.00 & 28.92 & 5522.05  & 82.50 & 29.50 & 64.24 & 4088.74  \\
 & DeBERTa   & 90.00 & 0.00  & 100.00 & 2763.66  & 77.50 & 53.50 & 30.97 & 5329.64  & 83.75 & 26.75 & 68.06 & 4046.65  \\
 & ChatGPT-Detector & 58.50 & 0.00  & 100.00 & 2606.75  & 93.00 & 73.00 & 21.51 & 5827.88  & 75.75 & 36.50 & 51.82 & 4217.32  \\
 & Flooding  & 87.50 & 0.00 & 100.00 & 2733.18 & 82.50 & 58.00 & 29.70 & 5447.84 & 85.00 & 29.00 & 65.88 & 4090.51  \\
 & RDrop & 95.00 & 10.00 & 89.47 & 3155.59 & 73.00 & 65.00 & 10.96 & 5973.84 & 84.00 & 37.50 & 55.36 & 4564.72 \\
 & RanMASK   & 67.00 & 2.00 & 97.01 & 2667.19 & 87.00 & 75.00 & 13.79 & 5433.82 & 77.00 & 38.50 & 50.00 & 4050.50  \\
 & RMLM      & 58.50 & 9.50  & 83.76  & 3397.99  & 92.00 & 72.50 & 21.20 & 5440.61  & 75.25 & 41.00 & 45.51 & 4419.30  \\
 \rowcolor{mygray}
 & SCRN      & 94.50 & 71.00 & 24.87  & 4419.16  & 70.50 & 54.50 & 22.70 & 5725.79  & 82.50 & 62.75 & 23.94 & 5072.48  \\
\midrule
\multirow{13}{*}{\rotatebox{90}{Deep-Word-Bug}} &
  Log-Likelihood &
  62.00 &
  0.00 &
  100.00 &
  352.02 &
  97.50 &
  97.50 &
  0.00 &
  1726.71 &
  79.75 &
  48.75 &
  38.87 &
  1039.36 \\
 & Log-Rank  & 64.50 & 0.00  & 100.00 & 354.59   & 97.50 & 97.00 & 0.51  & 1726.94  & 81.00 & 48.50 & 40.12 & 1040.77  \\
 & Entropy   & 77.50 & 0.00  & 100.00 & 351.23   & 74.00 & 67.00 & 9.46  & 1613.55  & 75.75 & 33.50 & 55.78 & 982.39   \\
 & GLTR      & 50.50 & 0.00  & 100.00 & 354.47   & 97.50 & 93.50 & 4.10  & 1666.85  & 74.00 & 46.75 & 36.82 & 1010.66  \\
 & SeqXGPT   & 85.50 & 0.00  & 100.00 & 353.94   & 88.00 & 88.00 & 0.00  & 1724.10  & 86.75 & 44.00 & 49.28 & 1039.02  \\
 & BERT      & 57.00 & 0.00  & 100.00 & 368.96   & 95.50 & 90.00 & 5.76  & 1675.84  & 76.25 & 45.00 & 40.98 & 1022.40  \\
 & RoBERTa   & 82.00 & 0.50  & 99.39  & 351.32   & 83.00 & 71.00 & 14.46 & 1464.67  & 82.50 & 35.75 & 56.67 & 908.00   \\
 & DeBERTa   & 90.00 & 1.50  & 98.33  & 353.25   & 77.50 & 64.50 & 16.77 & 1316.41  & 83.75 & 33.00 & 60.60 & 834.83   \\
 & ChatGPT-Detector & 58.50 & 0.50  & 99.15  & 346.69   & 93.00 & 87.00 & 6.45  & 1671.25  & 75.75 & 43.75 & 42.24 & 1008.97  \\
 & Flooding  & 87.50 & 4.00  & 95.43  & 359.58   & 82.50 & 68.00 & 17.58 & 1610.94  & 85.00 & 36.00 & 57.65 & 985.26   \\
 & RDrop & 95.00  & 16.50 & 82.63 & 514.99  & 73.00  & 62.00  & 15.07 & 1597.97 & 84.00  & 39.25 & 53.27 & 1056.48 \\
 & RanMASK   & 67.00 & 6.00 & 91.04 & 445.55 & 87.00 & 60.00 & 31.03 & 1353.78 & 77.00 & 33.00 & 57.14 & 899.66   \\
 & RMLM      & 58.50 & 3.00  & 94.87  & 354.36   & 92.00 & 66.00 & 28.26 & 1492.68  & 75.25 & 34.50 & 54.15 & 923.52   \\
 \rowcolor{mygray}
 & SCRN      & 94.50 & 50.00 & 47.09 & 1072.79 & 70.50 & 58.00 & 17.73 & 1535.02 & 82.50 & 54.00 & 34.55 & 1303.90  \\
\midrule
\multirow{13}{*}{\rotatebox{90}{Pruthi}} &
  Log-Likelihood &
  62.00 &
  0.00 &
  100.00 &
  48544.98 &
  97.50 &
  97.50 &
  0.00 &
  74465.43 &
  79.75 &
  48.75 &
  38.87 &
  61505.20 \\
 & Log-Rank  & 64.50 & 0.50  & 99.22  & 50907.28 & 97.50 & 95.50 & 2.05  & 74438.57 & 81.00 & 48.00 & 40.74 & 62672.92 \\
 & Entropy   & 77.50 & 0.00  & 100.00 & 37069.67 & 74.00 & 69.00 & 6.76  & 66872.37 & 75.75 & 34.50 & 54.46 & 51971.02 \\
 & GLTR      & 50.50 & 0.50  & 99.01  & 50694.88 & 97.50 & 95.00 & 2.56  & 74298.09 & 74.00 & 47.75 & 35.47 & 62496.48 \\
 & SeqXGPT   & 85.50 & 0.50  & 99.42  & 49312.26 & 88.00 & 88.00 & 0.00  & 73819.71 & 86.75 & 44.25 & 48.99 & 61565.98 \\
 & BERT      & 57.00 & 1.00  & 98.25  & 49171.64 & 95.50 & 87.50 & 8.37  & 64115.08 & 76.25 & 44.25 & 41.97 & 56643.36 \\
 & RoBERTa   & 82.00 & 1.50  & 98.17  & 37416.09 & 83.00 & 54.00 & 34.94 & 60541.50 & 82.50 & 27.75 & 66.36 & 48978.80 \\
 & DeBERTa   & 90.00 & 1.50  & 98.33  & 30985.48 & 77.50 & 42.00 & 45.81 & 43183.26 & 83.75 & 21.75 & 74.03 & 37084.37 \\
 & ChatGPT-Detector & 58.50 & 0.00  & 100.00 & 24919.15 & 93.00 & 68.00 & 26.88 & 72614.00 & 75.75 & 34.00 & 55.12 & 48766.58 \\
 & Flooding  & 87.50 & 4.50  & 94.86  & 45652.25 & 82.50 & 54.50 & 33.94 & 51702.81 & 85.00 & 29.50 & 65.29 & 48677.53 \\
 & RDrop & 95.00 & 24.00 & 74.74 & 64496.99 & 73.00 & 48.00 & 33.25 & 48799.05 & 84.00 & 36.00 & 57.14 & 56648.02 \\
 & RanMASK   & 67.00 & 8.00 & 88.06 & 45649.39 & 87.00 & 30.50 & 64.94 & 38997.11 & 77.00 & 19.25 & 75.00 & 42323.25 \\
 & RMLM      & 58.50 & 2.00  & 96.58  & 51019.47 & 92.00 & 87.50 & 4.89  & 62667.59 & 75.25 & 44.75 & 40.53 & 56843.53 \\
 \rowcolor{mygray}
 & SCRN      & 94.50 & 51.50 & 45.50  & 86841.90 & 70.50 & 49.50 & 29.79 & 39481.41 & 82.50 & 50.50 & 38.79 & 63161.66 \\
\bottomrule
\end{tabular}
}
\caption{Results of cross-genre AIGT detection under different attack methods.}
\label{tab:details-cross-genre-with-attack}
\end{table*}

\subsection{Details of Mixed-source AIGT Detection under Adversarial Perturbations} \label{sec:Details of Mixed-source AIGT Detection under Adversarial Perturbations}

Table \ref{tab:details-mixed-source-with-attack} shows the robustness performance of all compared AIGT detectors on \texttt{SeqXGPT-Bench} dataset under three adversarial attack methods. The results consistently demonstrate the superior robustness of SCRN in mixed-source AIGT detection.

\begin{table*}[!h]
\centering
\scalebox{0.7}{
\begin{tabular}{ll|cccc|cccc|cccc}
\toprule
 &
   &
  \multicolumn{4}{c|}{AI $\rightarrow$ Human} &
  \multicolumn{4}{c|}{Human $\rightarrow$ AI} &
  \multicolumn{4}{c}{Overall} \\
& Methods & OA $\uparrow$ & AUA $\uparrow$ & ASR $\downarrow$ & ANQ $\uparrow$ & OA $\uparrow$ & AUA $\uparrow$ & ASR $\downarrow$ & ANQ $\uparrow$ & OA $\uparrow$ & AUA $\uparrow$ & ASR $\downarrow$ & ANQ $\uparrow$ \\
\midrule
\multirow{13}{*}{\rotatebox{90}{PWWS}}
 & Log-Likelihood & 72.00 & 0.50  & 99.31  & 1281.86  & 62.00 & 53.50 & 13.71 & 1667.91  & 67.00 & 27.00 & 59.70 & 1474.88  \\
 & Log-Rank       & 73.50 & 0.50  & 99.32  & 1286.24  & 62.50 & 56.00 & 10.40 & 1697.20  & 68.00 & 28.25 & 58.46 & 1491.72  \\
 & Entropy        & 63.00 & 0.00  & 100.00 & 1239.29  & 55.50 & 27.50 & 50.45 & 1396.39  & 59.25 & 13.75 & 76.79 & 1317.84  \\
 & GLTR           & 76.50 & 0.00  & 100.00 & 1260.99  & 67.50 & 19.00 & 71.85 & 1285.64  & 72.00 & 9.50  & 86.81 & 1273.32  \\
 & SeqXGPT        & 96.50 & 65.00 & 32.64  & 1867.81  & 96.00 & 70.00 & 27.08 & 1893.98  & 96.25 & 67.50 & 29.87 & 1880.90  \\
 & BERT           & 90.50 & 1.00  & 98.90  & 1204.52  & 90.00 & 59.00 & 34.44 & 1815.44  & 90.25 & 30.00 & 66.76 & 1509.98  \\
 & RoBERTa        & 95.50 & 64.50 & 32.46  & 1840.19  & 93.00 & 62.50 & 32.80 & 1729.72  & 94.25 & 63.50 & 32.63 & 1784.96  \\
 & DeBERTa        & 95.50 & 54.50 & 42.93  & 1764.94  & 96.00 & 80.00 & 16.67 & 1940.47  & 95.75 & 67.25 & 29.77 & 1852.70  \\
 & Flooding       & 96.00 & 60.50 & 36.98  & 1800.01  & 95.50 & 53.00 & 44.50 & 1610.45  & 95.75 & 56.75 & 40.73 & 1705.23  \\
 & RDrop & 96.50 & 69.00 & 28.50 & 1819.95 & 95.00 & 70.00 & 26.32 & 1815.62 & 95.75 & 69.50 & 27.42 & 1817.78 \\
 & RanMASK        & 94.00 & 60.00 & 36.17  & 1784.11  & 86.00 & 71.00 & 17.44 & 1715.72  & 90.00 & 65.50 & 27.22 & 1749.92  \\
 & RMLM           & 91.00 & 69.00 & 24.18  & 1879.96  & 91.50 & 78.00 & 14.75 & 1986.50  & 91.25 & 73.50 & 19.45 & 1933.23  \\
 \rowcolor{mygray}
 & SCRN           & 95.00 & 87.00 & 8.42   & 1986.98  & 96.00 & 91.50 & 4.69  & 2099.91  & 95.50 & 89.25 & 6.54  & 2043.44  \\
\midrule
\multirow{13}{*}{\rotatebox{90}{Deep-Word-Bug}} 
 & Log-Likelihood & 72.00 & 0.50  & 99.31  & 162.17   & 62.00 & 60.50 & 2.42  & 535.48   & 67.00 & 30.50 & 54.48 & 348.82   \\
 & Log-Rank       & 73.50 & 0.50  & 99.32  & 163.11   & 62.50 & 60.00 & 4.00  & 539.27   & 68.00 & 30.25 & 55.51 & 351.19   \\
 & Entropy        & 63.00 & 0.50  & 99.21  & 156.12   & 55.50 & 46.50 & 16.22 & 456.98   & 59.25 & 23.50 & 60.34 & 306.55   \\
 & GLTR           & 76.50 & 0.00  & 100.00 & 158.01   & 67.50 & 41.00 & 39.26 & 385.61   & 72.00 & 20.50 & 71.53 & 271.81   \\
 & SeqXGPT        & 96.50 & 2.00  & 97.93  & 171.14   & 96.00 & 68.50 & 28.65 & 547.31   & 96.25 & 35.25 & 63.38 & 359.23    \\
 & BERT           & 90.50 & 7.50  & 91.71  & 180.22   & 90.00 & 76.50 & 15.00 & 515.24   & 90.25 & 42.00 & 53.46 & 347.73   \\
 & RoBERTa        & 95.50 & 71.00 & 25.65  & 373.37   & 93.00 & 50.50 & 45.70 & 379.60   & 94.25 & 60.75 & 35.54 & 376.48   \\
 & DeBERTa        & 95.50 & 69.50 & 27.23  & 233.89   & 96.00 & 82.00 & 14.58 & 440.42   & 95.75 & 75.75 & 20.89 & 337.16   \\
 & Flooding       & 96.00 & 70.00 & 27.08  & 361.76   & 95.50 & 54.50 & 42.93 & 394.68   & 95.75 & 62.25 & 34.99 & 378.22   \\
 & RDrop & 96.50 & 73.00 & 24.35 & 519.21 & 95.00 & 61.50 & 35.26 & 504.20 & 95.75 & 67.25 & 29.77 & 511.71  \\
 & RanMASK        & 94.00 & 57.50 & 38.83 & 464.99 & 86.00 & 75.00 & 12.79 & 622.30 & 90.00 & 66.25 & 26.39 & 543.64   \\
 & RMLM           & 91.00 & 73.50 & 19.23  & 428.07   & 91.50 & 73.00 & 20.21 & 541.52   & 91.25 & 73.25 & 19.73 & 484.79   \\
 \rowcolor{mygray}
 & SCRN           & 95.00 & 83.00 & 12.63  & 586.14   & 96.00 & 91.50 & 4.69  & 640.43   & 95.50 & 87.25 & 8.64  & 613.28   \\
\midrule
\multirow{13}{*}{\rotatebox{90}{Pruthi}}
 & Log-Likelihood & 72.00 & 0.50  & 99.31  & 15511.22 & 62.00 & 60.00 & 3.23  & 33247.44 & 67.00 & 30.25 & 54.85 & 24379.33 \\
 & Log-Rank       & 73.50 & 1.00  & 98.64  & 17032.86 & 62.50 & 59.50 & 4.80  & 33279.58 & 68.00 & 30.25 & 55.51 & 25156.22 \\
 & Entropy        & 63.00 & 0.50  & 99.21  & 11240.34 & 55.50 & 36.00 & 35.14 & 27676.25 & 59.25 & 18.25 & 69.20 & 19458.29 \\
 & GLTR           & 76.50 & 0.00  & 100.00 & 12173.33 & 67.50 & 21.00 & 68.89 & 22492.95 & 72.00 & 10.50 & 85.42 & 17333.14 \\
 & SeqXGPT        & 96.50 & 1.00 & 98.96 & 18296.61 & 96.00 & 64.50 & 32.81 & 32981.87 & 96.25 & 32.75 & 65.97 & 25639.24 \\
 & BERT           & 90.50 & 1.00  & 98.90  & 14234.70 & 90.00 & 67.50 & 25.00 & 34276.58 & 90.25 & 34.25 & 62.05 & 24255.64 \\
 & RoBERTa        & 95.50 & 72.00 & 24.61  & 33905.64 & 93.00 & 37.00 & 60.22 & 19640.80 & 94.25 & 54.50 & 42.18 & 26773.22 \\
 & DeBERTa        & 95.50 & 72.50 & 24.08  & 34325.91 & 96.00 & 71.00 & 26.04 & 32058.96 & 95.75 & 71.75 & 25.07 & 33192.44 \\
 & Flooding       & 96.00 & 68.00 & 29.17  & 32915.51 & 95.50 & 30.00 & 68.59 & 20405.92 & 95.75 & 49.00 & 48.83 & 26660.72 \\
 & RDrop & 96.50  & 70.50 & 26.94 & 32959.75 & 95.00  & 51.50  & 45.79 & 25669.62 & 95.75  & 61.00 & 36.29 & 29314.68 \\
 & RanMASK        & 94.00 & 52.00 & 44.68 & 30767.48 & 86.00 & 58.00 & 32.56 & 25411.29 & 90.00 & 55.00 & 38.89 & 28089.38 \\
 & RMLM           & 91.00 & 70.00 & 23.08  & 33542.23 & 91.50 & 69.50 & 24.04 & 32213.99 & 91.25 & 69.75 & 23.56 & 32878.11 \\
 \rowcolor{mygray}
 & SCRN           & 94.50 & 80.50 & 14.81  & 36077.57 & 95.50 & 78.50 & 17.80 & 37299.11 & 95.00 & 79.50 & 16.32 & 36688.34 \\
\bottomrule
\end{tabular}
}
\caption{Results of mixed-source AIGT detection under different attack methods.}
\label{tab:details-mixed-source-with-attack}
\end{table*}

\subsection{AI Assistant Statement}
Following the ACL 2023 Policy on AI Writing Assistance, we use AI assistant purely for the language of the paper, containing spell-checking, grammar-checking, and polishing our original content without suggesting new content. We affirm that all words refined by the AI assistant have been carefully reviewed and either rechecked or modified by us.

\end{document}